\documentclass[10pt,journal,compsoc]{IEEEtran}
\ifCLASSOPTIONcompsoc
  \usepackage[nocompress]{cite}
\else
  \usepackage{cite}
\fi
\usepackage{comment,verbatim}
\usepackage{amssymb, amsmath, amsthm, amsfonts}
\usepackage{tikz}
\usetikzlibrary{shapes, arrows, positioning}
\newcommand{\norm}[1]{\left\lVert#1\right\rVert}
\newtheorem{t1}{Theorem}
\usepackage{hyperref}
\usepackage{algorithm}
\usepackage{algorithmic}
\usepackage{adjustbox}
\usepackage{epsfig}
 \usepackage{color,soul}
\usepackage{subcaption}
\graphicspath{ {./Newfigures/} }
\renewcommand{\bar}{\overline}

\newcommand\copyrighttext{%
\footnotesize This work has been submitted to the IEEE for possible publication. Copyright may be transferred without notice, after which this version may no longer be accessible.}
\newcommand\copyrightnotice{%
\begin{tikzpicture}[remember picture,overlay]
\node[anchor=south,yshift=10pt] at (current page.south) {\fbox{\parbox{\dimexpr\textwidth-\fboxsep-\fboxrule\relax}{\copyrighttext}}};
\end{tikzpicture}%
}

\begin{document}
\title{Expectation Distance-based Distributional Clustering for Noise-Robustness}
\author{Rahmat Adesunkanmi,~\IEEEmembership{Student member,~IEEE},  Ratnesh~Kumar,~\IEEEmembership{Fellow,~IEEE}
%\vspace{-0.2in}
\thanks{This work was supported in part by U.S. National Science Foundation under grant NSF-CSSI-2004766 and NSF-PFI-2141084. The authors would like to thank  CoAgMET (Colorado's Mesonet) for access to many years of real-life weather data. }
\thanks{Rahmat Adesunkanmi, Ph.D. student, and Ratnesh Kumar, Palmer Professor, are both with the Department of Electrical and Computer
Engineering, Iowa State University, Ames, IA 50010 USA (e-mail:
rahma,rkumar@iastate.edu).}}

\maketitle
\copyrightnotice
\begin{abstract}
This paper presents a clustering technique that reduces the susceptibility to data noise by learning and clustering the data-distribution and then assigning the data to the cluster of its distribution. In the process, it reduces the impact of noise on clustering results. This method involves introducing a new distance among distributions, namely the expectation distance (denoted, ED), that goes beyond the state-of-art distribution distance of optimal mass transport (denoted, $W_2$ for $2$-Wasserstein): The latter essentially depends only on the marginal distributions while the former also employs the information about the joint distributions. Using the ED, the paper extends the classical $K$-means and $K$-medoids clustering to those over data-distributions (rather than raw-data) and introduces $K$-medoids using $W_2$. The paper also presents the closed-form expressions of the $W_2$ and ED distance measures. The implementation results of the proposed ED and the $W_2$ distance measures to cluster real-world weather data as well as stock data are also presented, which involves efficiently extracting and using the underlying data distributions---Gaussians for weather data versus lognormals for stock data. The results show striking performance improvement over classical clustering of raw-data, with higher accuracy realized for ED. Also, not only does the distribution-based clustering offer higher accuracy, but it also lowers the computation time due to reduced time-complexity.
\end{abstract}

\section{Introduction}
\IEEEPARstart{C}LUSTERING, a widely studied unsupervised learning technique, is commonly used in many fields for data analysis to make valuable inferences by observing what group each data point falls into. Classical clustering methods of $K$-means and $K$-medoids iteratively group raw-data into "similarity classes" depending on their relative distances or similarities. Optimal clustering aims to group the data points into clusters so that the data's total distance to their assigned cluster centers is minimized \cite{5467074}. 

The classical clustering algorithms work with raw-data and are not designed to be robust to uncertain/noisy data. However, data is naturally and inherently affected by the random nature of the physical generation process and measurement inaccuracies, sampling discrepancy, outdated data sources, or other errors, making it prone to noise/uncertainty \cite{ Chau2005UncertainDM, agg, aggarwal2009managing}. As an application, consider a weather station that monitors and measures daily the variables like temperature, humidity, and vapor pressure. The data is naturally noisy due to physical measurement equipment (that introduces thermal noise) and variations resulting from other natural sources such as lightning and solar flare. While the daily weather conditions are expected to be within specific predicted ranges for certain seasons, there can be days when those will vary. 

Clustering uncertain data has been well-recognized as a challenge in the data mining fields \cite{6051435}, having applications in diverse fields such as weather forecasting, medical diagnosis, image processing, and many more. Addressing the uncertainties in data can significantly improve the accuracy and robustness of clustering results. In the presence of data uncertainty, accounting for noise distribution and its impact on data values toward data clustering is meaningful, and so one viable way to reduce the impact of uncertainty in data clustering is to extract and utilize its probability distribution whenever feasible. Accordingly, \cite{6051435, 10.1145/872757.872823, Pei2007ProbabilisticSO, 10.5555/1083592.1083699} cluster the data-distributions, estimated from datasets belonging to the same random variables, and assign raw-data to its distribution cluster. 

\subsection{Our Contributions}
This paper studies the clustering of uncertain/noisy data, making the following contributions:
\begin{itemize}
\item For clustering data distributions, we propose a new distance measure among distributions, namely Expectation Distance (ED), which extends the widely used $2$-Wasserstein ($W_2$) distance by factoring in the correlation information ignored by $W_2$. We formally show that ED meets all the required criteria of being a metric. The proposed ED measure can also be utilized in other machine-learning applications (not just clustering) where distributional distances can yield noise-robust results.
\item We provide a closed-form formula for the $W_2$ and ED distances in terms of the means and covariances of the distributions and provide those parameters for Gaussian and lognormal distributions.
\item Using the proposed ED distance, we extend the classical clustering techniques of $K$-means and $K$-medoids to cluster over the data-distributions. We denote the corresponding $K$-means and $K$-medoids as EKM and EKMd, respectively. The corresponding $W_2$-distance-based versions are termed WKM and WKMd, respectively.
\item We show that the Barycenter is the cluster-center in both WKM and EKM.
In contrast, we show that the same is not true for WKMd vs. EKMd, which generally can have different cluster-center. 
\item We provide time-complexity for clustering based on raw-data versus the distributions. It is shown that the latter is of lower complexity, yet it offers higher accuracy.
\item We implement and compare the results of all six clustering algorithms: Classical $K$-means and $K$-medoids for raw-data vs. $W_2$-based vs. ED-based clustering of data distributions, by applying to real-world noisy weather data with Gaussian characteristic and real-world stocks data having lognormal distribution.  
\end{itemize}

\subsection{Related Works}
The task of distribution clustering requires distance metrics among the distributions. The Maximum Mean Discrepancy (MMD) distance \cite{Gretton2012AKT} measures the difference between the mean of the probability distributions, while the Integral Probability Metrics (IPMs) \cite{Muller1997IntegralPM} measures the distance based on the integrals of a discrepancy function. MMD is restrictive by being limited to only the mean values, and IPMs are generally computationally expensive.
Optimal mass transport (OMT) is a commonly used metric that seeks to find the least costly way of transforming one distribution of mass to another relative to a given transport cost \cite{7974883}. The OMT has been increasingly used in recent years in various applied fields such as economics \cite{galichon2014stochastic}, image processing \cite{rabin2011wasserstein}, machine learning \cite{blanchet2019robust}, data science \cite{peyre2019computational}, among others. $W_2$-distance or 2-Wasserstein distance \cite{givens1984class}  uses the OMT concept where the cost of transportation is the expectation of the Euclidean distance. $W_2$-distance has been used in clustering algorithms, such as Wasserstein $K$-means \cite{Domazakis2019ClusteringMD, Ye2017FastDD, verdinelli2019hybrid}, and also as a Wasserstein auto-encoder \cite{tolstikhin2017wasserstein}. However, the Wasserstein distance only considers the pairwise marginal distribution information and ignores the true correlation information. This paper proposes a new distance metric, the Expectation Distance (ED), that can account for the uncertainty and factor in the correlation information.
 
Classical clustering algorithms often require complete data. Several techniques have been proposed to handle missing data in clustering: Imputation-based methods \cite{GAIN, KNNImpute} fill the missing values with estimated values, while subspace-based methods \cite{EMimpute, kriegel2009clustering} identify subspaces with no missing values and cluster the data in those subspaces. A probabilistic method in \cite{MCimpute} models the distribution of missing values and uses that model to generate the missing data. 

\subsection{Clustering Definition}
Consider a set, $S$, that needs to be clustered into $K$ number of clusters. The clustering problem requires finding a function, $C:S \xrightarrow{}[1,K]$, to map elements of $S$ to one of the $K$ clusters in some optimal sense. Then for each $i\in[1,K]$, the $i^{th}$ cluster set under the clustering $C$ is given by,
\begin{equation}
 S_C(i) := \{s \in S|C(s) = i\},   
\end{equation}
and its cluster-center is the minimizer of the distance to the cluster members:
\begin{align}\label{eq:omtc}
  \bar{s}_C(i) := \arg \left\{ \min_{s} \sum_{s'\in S_C(i)} \norm{s - s'}_2 \right\},\forall i\in[1,K],
\end{align}
where the notation $|.|$ measures the size of its argument set.
The cluster-center turns out to be the center-of-mass, also called the Barycenter, of the cluster members: 
\begin{equation}\label{eq:bc}
\bar{s}_C(i) := \frac{\sum_{s \in S_C(i)}  s}{|S_C(i)|},\forall i\in[1,K].
\end{equation} 

The goal of clustering is to find an optimal $K$-cluster that minimizes the aggregate distances of each of the data to their respective cluster-centers, i.e.,
\begin{equation}
\begin{aligned}
  \min_C \sum_{i=1}^K \!\sum_{s' \in S_C(i)} \!\!\!\!\norm{\bar{s}_C(i)-s'}_2  & \\ 
  =\min_C \sum_{i=1}^K \!\sum_{s' \in S_C(i)} \!\norm{\frac{\sum_{s \in S_C(i)}  s}{|S_C(i)|} \!-\!s'}_2\!.
\end{aligned}
\end{equation}
 The corresponding optimal cluster is called $K$-means.

For $K$-means, a cluster-center is a Barycenter and may not coincide with any of the data points. If we require the cluster-center be one of the data points, then the resulting clustering is called $K$-medoids for which the objective function can be written as:
\begin{align}
\min_C \sum_{i=1}^K \left\{ \min_{s \in S_C(i)} \left( \sum_{s' \in S_C(i)} \norm{s-s'}_2\right ) \right\}.
\end{align}
Here the inner optimization minimizes the distance between one data point in a cluster to all other data points within the same cluster to determine a cluster-center:
\begin{align}\label{eq:edc}
  \hat{s}_C(i) := \arg \left\{ \min_{s \in S_C(i)} \sum_{s'\in S_C(i)}\norm{s - s'}_2 \right\},\forall i\in[1,K].
\end{align}
One popular heuristic to find a locally optimal clustering involves starting with an arbitrary initial clustering, $C_0$, and iteratively finding a better clustering $C_{n+1}$ from a prior clustering $C_n, (n\geq 0)$, until this process converges, i.e., until $C_{n+1}= C_n$. The heuristic finds the $i^{th}$ cluster of the $(n + 1)^{th}$ iteration as the set of those elements that are nearest to the $i^{th}$ cluster-center of the $n^{th}$ iteration. The same iterative computation for $K$-means can be used to find $K$-medoids with the change that the cluster-center is restricted to a data point.

 \section{clustering using Data-Distributions for Noise-Robustness}
One approach to extend the $K$-Means and $K$-medoids and make them robust to noise-led outliers is to perform clustering over the data-distributions and then assign each raw-data to the cluster of its distribution. This way, the effect of outliers is reduced, making the clustering more robust. Clustering over data-distributions requires measuring distances between distribution pairs, and for this, we present a new "Expectation Distance" (ED) and also utilize the commonly used Optimal Mass Transport (OMT) distance, also called $W_2$ distance for comparison.

\subsection{Optimal Mass Transport / \texorpdfstring{$W_2$}{}-Distance}
OMT computes the distance between two random variables $X$ and $Y$ having distributions $f_X$ and $f_Y$, respectively, by associating cost to "transport" the probability mass from the starting distribution $f_X$ to the destination distribution $f_Y$, while minimizing that cost among all possible transports. Letting $T\;:\;{\mathbb R}^{n}\to {\mathbb R}^{n}$ denote a transport map, OMT minimizes the associated cost of transport:
\begin{equation} 
\min_T\int _{\mathbb R^{n}} c(x,T(x))f_X(dx),
\end{equation}
where $c(\cdot,\cdot)$ is a user-specified cost function of transport. Kantorovich proposed the cost to be Euclidean distance and minimized the transport cost over the joint distributions $f_{XY}$ so that the marginals along the two coordinate directions coincide with $f_X$ and $f_Y$, respectively, resulting in the $2$-Wasserstein or $W_2$-distance\cite{givens1984class}:
\begin{equation}\label{j}
W_2^2(X,Y)\!:=\!\!\!\!\!\!\!\!\!\!\!\!\!\! \inf_{\tiny\begin{array}{c}f_{XY}:\\ E_X(Y|X)=f_Y,\\E_Y(X|Y)=f_X\end{array}}\!\!\!\!\int _{\mathbb R^{n}\times {\mathbb R} ^{n}}\norm{x-y}_2^{2}f_{XY}(x,y)dxdy.\end{equation}

\subsubsection{Formula for \texorpdfstring{$W_2$}{}}
For a random $X$, we let $\mu_X:=\mathbb E(X)$ denote the mean of $X$, similarly for another random variable $Y$, $\mu_Y:=\mathbb E(Y)$ is its mean, and their covariance is denoted $\Sigma_{XY}:=\mathbb E[(X-\mu_X)(Y-\mu_Y)^T]$. The variances of $X$ and $Y$ are denoted $\Sigma_X:=\Sigma_{XX}$ and $\Sigma_Y:=\Sigma_{YY}$ respectively. To compute $W_2(X,Y)$, consider the term in (\ref{j}) that needs to be minimized:
\begin{equation}\label{ee} \begin{aligned}
&\int _{\mathbb R^{n}\times {\mathbb R} ^{n}}\norm{x-y}_2^{2}f_{XY}(x,y)dxdy\\
&={\mathbb E}[\norm{X-Y}_2^{2}]  \\
&={\mathbb E}[\norm{(X-\mu_X+\mu_X)\!-\!(Y-\mu_Y+\mu_Y)}_2^{2}]\\
&={\mathbb E}[\norm{(X-\mu_X)-(Y-\mu_Y)}_2^{2}]\!+\!\norm{\mu_{X}\!-\!\mu_{Y}}_2^{2}\\
&=\mathrm{trace}(\Sigma _{X}+\Sigma _{Y}-2\Sigma_{XY})\!+\!\norm{\mu_{X}\!-\!\mu_{Y}}_2^{2}.
\end{aligned} 
\end{equation}
Note that ${\mathbb E}[\norm{X-Y}_2^{2}]$ only depends on the first two moments---This is because the 2-norm is used for measuring the distance. If instead $p$-norm, $p>2$, is used, then higher-order moments will be required.

For computing $W_2(X,Y)$, we need to minimize (\ref{ee}) with respect to those joint distributions $f_{XY}$ that possess the marginals $f_X$ and $f_Y$. Fixing the marginals $f_X$ and $f_Y$ fixes $\mu_X,\mu_Y,\Sigma_X,\Sigma_Y$, leaving $\Sigma_{XY}$ to be the only variable of optimization. Since (\ref{ee}) is a decreasing function of $\Sigma_{XY}$, it is then obvious that the minimization will be achieved when $\Sigma_{XY}$ is the largest, i.e., $X$ and $Y$ are the most correlated. Mathematically, we need to solve the following semidefinite program :
 \begin{equation}\label{eq:w2min} 
 \begin{aligned} 
 &\min _{\Sigma_{XY}}\left[ \mathrm{trace}(\Sigma _{X}+\Sigma _{Y}-2\Sigma_{XY})+\norm{\mu_{X}-\mu_{Y}}_2^{2}\right]\\
 &\mbox{s.t. } { \left [{\begin{matrix} \Sigma _{X} & \Sigma_{XY} \\ \Sigma_{XY}^T & \Sigma _{Y} \end{matrix}}\right]}\ge 0. \end{aligned}\end{equation}
The minimum in (\ref{eq:w2min}) is achieved at :
 \begin{equation}\label{eq:sigmamin}
 \Sigma_{XY}= (\Sigma _{X}^{1/2}\Sigma _{Y}\Sigma _{X}^{1/2})^{1/2}.\end{equation}
Thus the $W_2$ distance has the closed-form formula:
\begin{equation}\label{eq:omt}
\begin{aligned}
    W_2^2 (X, Y)= \norm{\mu_X -\mu_Y}_2^2 \\ 
    +\, \mathrm{trace}[\Sigma_X +\Sigma_Y - 2(\Sigma_X^{\frac{1}{2}}\Sigma_Y\Sigma_X^{\frac{1}{2}})^{\frac{1}{2}}].
\end{aligned}
\end{equation}
Several numerical methods have been developed to compute the Wasserstein distance efficiently, such as the Sinkhorn algorithm \cite{Cuturi2013Sinkhorn} and the Entropic Regularization of Optimal Transport (EROT) \cite{Frogner2015LearningWT}.

\subsubsection{Cluster-center under \texorpdfstring{$W_2$}{}}
The cluster-center for a cluster set $S$ of distributions in the case of $W_2$-based $K$-means, denoted WKM, is given by:
\begin{equation}\label{eq:baryW}
  \arg \min_{X'} \sum_{X\in S} W_2^2(X',X).
\end{equation}
The cluster-center turns out to be the Barycenter \cite{Domazakis2019ClusteringMD}:
\begin{equation}\label{orig:bary}
\frac{1}{|S|}\sum_{X\in S}X. 
\end{equation}
In contrast, in the case of the $W_2$-based $K$-medoids, denoted WKMd, a cluster-center is restricted to be chosen from one of the data points and may differ from the Barycenter:
\begin{equation}
  \arg \min_{X'\in S} \sum_{X\in S} W_2^2(X',X).
\end{equation}
 For distributions $\{X_i, 1\leq i\leq n\}$, with $\mathbb E(X_i)=\mu_i, Var(X_i)=\Sigma_i$, the Barycenter distribution's mean $\mu$ and covariance $\Sigma$ are given by :
\begin{equation} \label{msig}
    \mu= \frac{1}{n}\sum_{i=1}^n \mu_i, \text{   and   }
  \Sigma   = \frac{1}{n}\sum_{i=1}^n (\Sigma^{\frac{1}{2}} \Sigma_i\\\Sigma^{\frac{1}{2}})^{\frac{1}{2}}.
\end{equation}	
(\ref{msig}) provides $\Sigma$ in an implicit form, and its computation is a fixed point of the following iteration \cite{ALVAREZESTEBAN2016744}:

\begin{equation}\label{sig}
\Sigma_{n+1} = \Sigma_n ^{-\frac{1}{2}}\left (\frac{1}{N}{\sum _{i=1}^{N}  (\Sigma_n ^{\frac{1}{2}} \Sigma_i\Sigma_n ^{\frac{1}{2}})^{\frac{1}{2}}}\right)^{2}\Sigma_n ^{-\frac{1}{2}}.\end{equation}

\subsection{Expectation Distance}
While $W_2$-based distance measure is popular, it ignores the true correlation information: The minimization in (\ref{eq:w2min}) is achieved when the two given marginals are most correlated, which may not be the case. Recognizing this limitation of $W_2$ distance, we hereby propose a new and more general distance measure between any two probability distributions that also accounts for their joint distributions (and not just their marginals); it is simply the expectation distance (ED) of the given random variables $X$ and $Y$: 
\begin{equation}\label{eq:ed}
d^2_{X,Y}:=E[\norm{X-Y}_2^2]= \int _{\mathbb R^{n}\times {\mathbb R} ^{n}}\norm{x-y}_2^{2}f_{XY}(x,y)dxdy.
\end{equation}

The following result establishes that the above definition provides a metric over the distributions.
\begin{t1}\rm
$d_{X, Y} = \left[ E[\norm{X-Y}_2^2]\right]^{\frac{1}{2}}$ in definition (\ref{eq:ed}) meets all the required criteria of being a distance measure (namely, positivity, symmetry, zero if and only if equal, and triangular inequality).
\end{t1}
\noindent{\bf Proof:} The proof below takes into consideration some common properties of expected value and the fact that $\|x-y\|$ is a metric in itself (and satisfies the said four properties).
\begin{description}
\item[Positivity ($d_{X,Y} \geq 0$):]\mbox{}\\
If a random variable $p$ is non-negative, its expected value is also non-negative, i.e., $p \geq 0 \Longleftrightarrow  E(p)\geq 0$. Then since
$p := \norm{x-y}_2^2 \geq 0$, it holds that $d^2_{X,Y} \geq 0  \Longleftrightarrow d_{X,Y} \geq 0$ ($d_{X,Y}$ being the positive square root of $d_{X,Y}^2$). 

\item[Symmetry ($d_{X,Y}  = d_{Y,X}$):]\mbox{}\\
 We have $p_{x,y} := \norm{x-y}_2^2= \norm{y-x}_2^2 =:p_{y,x}$. Then $p_{x,y}  = p_{y,x} \implies E[p_{x,y}]=E[p_{y,x}] \implies d_{X,Y}^2=d_{Y,X}^2\implies d_{X,Y}=d_{Y,X}$.
 
\item[Zero iff equal ($d_{X,Y}  = 0 \implies X=Y$):]\mbox{}\\
 The non-degeneracy property of an expected value asserts that $p = 0  \Leftrightarrow E[p] = 0$ for equivalence classes of almost surely equal variables. Then $ X = Y \Leftrightarrow X-Y=0 \Leftrightarrow p:=\norm{x-y}_2^2 = 0 \Leftrightarrow  E[p] = 0  \Leftrightarrow d_{X,Y}^2 = 0 \Leftrightarrow d_{X,Y}=0$.  

\item[Triangle Inequality ($d_{X,Z} \leq  d_{X,Y} + d_{Y,Z}$):]\mbox{}\\
 By Minkowski inequality in $L^{\mathbb{P}}$ spaces:\\
  $d_{X,Z}=\left[ E[\norm{X-Z}_2^2]\right]^{\frac{1}{2}}$\\
  $\leq \left[ E[\norm{X-Y}_2 + \norm{Y-Z}_2]^2\right]^{\frac{1}{2}}$\\
  $\leq \left[ E[\norm{X-Y}_2]^2\right]^{\frac{1}{2}} + \left[ E[\norm{Y-Z}_2]^2\right]^{\frac{1}{2}}$\\
  $=  d_{X,Y} + d_{Y,Z}$.
\end{description}
These properties conclude that the ED proposed in (\ref{eq:ed}) is a distance measure over distributions. \qed

\subsubsection{Formula for ED}
Given random variables $X,Y,$ it follows
from the definition (\ref{eq:ed}) and equality (\ref{ee}) that their ED is given by:
\begin{align}\label{ex} 
&d_{X,Y}^2={\mathbb E}{\norm{X-Y}_2^2}\nonumber\\ 
&= \mathrm{trace} (\Sigma_X  + \Sigma_Y  -2 \Sigma_{XY} ) + \norm{\mu_X-\mu_Y}_2^2.
\end{align}
It can be noted that whenever the correlation $\Sigma_{XY}$ of the two random variables is the same as the one given in (\ref{eq:sigmamin}), the ED distance coincides with the $W_2$ distance. However, in general, ED is of a higher value: To attain the minimization of (\ref{eq:w2min}), which is a decreasing function of $\Sigma_{XY}$, its largest possible value (i.e., most correlated) gets picked, but in reality, $\Sigma_{XY}$ may be smaller (i.e., less correlated), leading to ED being larger than $W_2$.

\subsubsection{Cluster-center under ED} 
\begin{t1}\rm
The cluster-center for EKM (ED-based $K$-means) is the Barycenter of the cluster-set (as in the case of $W_2$-based $K$-means).
\end{t1}
\noindent{\bf Proof:} Under the ED measure, the cluster-center in case of $K$-means for a cluster-set $S=\{X_1,\ldots,X_{|S|}\}$ of distributions is given by the following expression that optimizes the total distance between a candidate cluster-center distribution and each of the distributions in $S$, with respect to all possible choices for the cluster-center candidate distribution $X$, along with all possible choices for the joint distribution candidates between the candidate cluster-center and the elements of the cluster, $\{f_{X'X_i},1\leq i\leq|S|\}\mbox{ with }\forall i:E_{X'}(X_i|X')=X_i,E_{X_i}(X' |X_i)=X'$:
\begin{align}\label{eq:edcc}
  &\arg\;\min_{X'} 
  \inf_{\tiny\begin{array}{c}f_{X'X_i},1\leq i\leq |S|:\\ E_{X'}(X_i|X')=X_i,\\E_{X_i}(X'|X_i)=X'\end{array}}  \sum_{1\leq i\leq |S|}{\mathbb E}\norm{X'-X_i}_2^2.
 \end{align}
Since the joint distributions ($f_{X'X_i}$ vs. $f_{X'X_j}, 1\leq i\neq j\leq |S|$) can be chosen independent of each other, and the minimization is of the sum of positive entries, the operations in Eq. (\ref{eq:edcc}) can be rearranged to obtain:
\begin{align*}
  &\arg\;\min_{X'} \sum_{1\leq i\leq |S|}
  \inf_{\tiny\begin{array}{c}f_{X'X_i},1\leq i\leq |S|:\\ E_{X'}(X_i|X')=X_i,\\ E_{X_i}(X'|X_i)=X'\end{array}}  {\mathbb E}\norm{X'-X_i}_2^2,\\
  &=\arg\min_{X'} \sum_{1\leq i\leq |S|} W_2^2(X',X_i),
\end{align*}
where the last equality follows from the definition of $W_2$-distance. It can then be seen that the last expression is the same as that of $W_2$-based distance in (\ref{eq:baryW}), and hence, the resulting cluster-center in the case of EKM is again the Barycenter of Eq.~(\ref{orig:bary}). \qed

In the case of $K$-medoids using ED distance, denoted EKMd, a cluster-center is chosen to be one of the data points, so their joint distribution, as already estimated from the dataset, is known and used to perform the optimization:
\begin{equation}
  \arg \min_{X'\in S} \sum_{X\in S} {\mathbb E}\norm{X'-X}_2^2.
\end{equation}
% Next, we discuss the value of ED for the case of Gaussian distributions.

 \subsection{Covariances with cluster-centers}
In the case of WKMd or EKMd, the cluster-center is one of the data points, so its joint distribution with any other data point, and hence the corresponding covariance, is already known. However, in the case of WKM or EKM, a cluster-center is the Barycenter of the cluster-set. We can compute its covariance with the other data points within its cluster-set, say, $\{X_1, X_2, \ldots, X_n\}$ as follows. The covariance between a data point $X_i$ and a cluster-center $X=\frac{X_1+X_2+\ldots+X_n}{n}$ is:
\begin{equation}\label{joint}
    \begin{aligned}
    &\Sigma_{X_iX} = Cov\left(X_i, \frac{1}{n}\sum_{j=1}^n X_j\right)\\
    &= \frac{1}{n}\sum_{j=1}^n \Sigma_{X_iX_j}.
\end{aligned}
\end{equation}
Eq.~(\ref{joint}) provides the closed-form expression to compute the covariance of the joint distribution between a data point and its cluster center in the case of WKM or EKM. In the case of EKM, since the pairwise joint distributions appearing in (\ref{joint}) are already known and fixed, those pairwise covariances are also known and fixed, and so for the case of EKM, (\ref{joint}) provides the final answer. However, in the case of WKM, (\ref{eq:sigmamin}) provides the optimum covariance between a pair of distributions, and hence for the case of WKM, the covariance between a cluster-element and its cluster-center is given by,
\[\Sigma_{X_iX}=\frac{1}{n}\sum_{j=1}^n \left(\Sigma_{X_i}^{1/2}\Sigma_{X_j}\Sigma_{X_i}^{1/2}\right)^{1/2}.\]

\subsection{\texorpdfstring{$W_2$}{} and ED for Lognormals}\label{lne}
From the definition of the lognormal distribution, it is known that if $X\sim lognormal(\theta_X,\Delta_X)$ and $Y\sim lognormal(\theta_Y,\Delta_Y)$ are multivariate lognormal random variables with parameters $(\theta_X,\Delta_X)$ and $(\theta_Y,\Delta_Y)$, respectively, then $A=[A(i) := \ln(X(i))]_{n\times 1}$ and $B=[B(i) := \ln(Y(i))]_{n\times 1}$, are multivariate normal random variables with $A\sim\mathcal N(\theta_X,\Delta_X),B\sim\mathcal N(\theta_X,\Delta_X)$. Then the means and covariances of the two lognormal random variables are as given:
\begin{align*}
&\mu_X = \mathbb E[X]=[\exp(\theta_X(i)+0.5\Delta_X(ii))]_{n\times 1}\\
&\mu_Y = \mathbb E[Y]=[\exp(\theta_Y(i)+0.5\Delta_Y(ii))]_{n\times 1}\\
&\mathbb E[XY^T]= \mathbb E[YX^T]=[\exp(\theta_X(i) + \theta_Y(j)\\
&\quad\quad\quad\quad\quad+0.5(\Delta_X(ii)
+ \Delta_Y(jj) + 2\Delta_{XY}(ij)))]_{n\times n} \\
&\Sigma_{XY}=\mathbb E(XY^T)-\mathbb E(X)\mathbb E(Y)^T\\
&\quad=[(\exp(\theta_X(i)+.5\Delta_X(ii))\times((\exp(\theta_Y(j)+.5\Delta_Y(jj))\\
&\qquad\times(\exp(\Delta_{XY}(ij))-1)]_{n\times n}\\
&\Sigma_X=\Sigma_{XX};\quad \Sigma_Y=\Sigma_{YY}.
\end{align*}

To compute the $W_2$ and ED measures for lognormal distributions, we can plug the above parameters for the lognormals into the formulas for $W_2$ (\ref{eq:omt})  and ED (\ref{ex}) respectively, to get the two respective distances.

\section{\texorpdfstring{$W_2$}{}- \& ED-based distribution clustering }
Here we extend the classical $K$-means and $K$-medoids-based clustering methods to clustering over the data-distributions (as opposed to raw-data). The distance measures considered in clustering over the data-distributions are the above-mentioned $W_2$ and ED distances. The corresponding WKM and EKM clustering algorithms are presented in Algorithms \ref{algm1} and \ref{algm2}, respectively,  and the corresponding WKMd and EKMd clustering algorithms are presented in Algorithms \ref{algm3} and \ref{algm4} respectively. Each algorithm starts with an initial guess of $K$ cluster-centers, iteratively assigns data to its nearest cluster-center, then recomputes the cluster-centers and repeats until convergence.

\begin{algorithm}
\caption{Distributional $K$-means using $W_2$ (WKM)} \label{algm1}
\begin{algorithmic}[1]
\REQUIRE N distributions, $f_i\sim f_1,f_2, \ldots,f_{N}$
  \STATE Choose $K$ initial cluster-centers ${f_{c_1}, f_{c_2}, \ldots, f_{c_k}}$ from the given set of $N$ distribution data.
 \FOR {$i = 1$ to $N (=${\rm total number of distributions}) }
  \STATE Solve \\
  $k_i=\arg \left\{ \displaystyle \min_{1\leq k\leq K} W_2^2(f_i, f_{c_k}) \right\}$ \\
  \STATE Assign $f_i$ to cluster $k_i$
 \ENDFOR\\
 \texttt{This creates a disjoint partition of the data into subsets $f_1, f_2,\ldots,f_K$.}
 \FOR {$k = 1$ to $K (=${\rm total number of clusters}) }
  \STATE Update center $f_{c_k} = \mathit{Barycenter}(f_k)$
 \ENDFOR\\
 \texttt{Repeat  steps 2 to 8 using new $c_k$'s until convergence.}\\
 \STATE Group data points using the final  distribution groups in $f_1, f_2,\ldots,f_K$
 \end{algorithmic}
\end{algorithm}
\pagebreak
\begin{algorithm}
\caption{Distributional $K$-means using ED (EKM)} \label{algm2}
\begin{algorithmic}[3]
\REQUIRE N distributions, $f_i\sim f_1,f_2, \ldots,f_{N}$
\STATE Steps same as Algorithm 1, with the following changed step~3:
\newline
  \hspace*{-.15in}3:$\qquad k_i=\arg \left\{\displaystyle \min_{1\leq k\leq K}  d_{f_i,f_{c_k}}^2\right\}$.  
 \end{algorithmic}
\end{algorithm}
\begin{algorithm}
\caption{Distributional $K$-medoids using $W_2$ (WKMd)} \label{algm3}
 \begin{algorithmic}[1]
 \REQUIRE N distributions, $f_i\sim f_1,f_2, \ldots,f_{N}$
 \STATE Choose $K$ initial cluster-medoids ${f_{c_k}, f_{c_k}, \ldots, f_{c_k}}$ from among the given set of $N$ data, $X=\{f_1,f_2,\ldots,f_N\}$.
 \FOR {$i = 1$ to $N (=${\rm total number of distributions}) }
  \STATE Solve \\
  $k_i=\arg \left\{ \displaystyle \min_{1\leq k\leq K} W_2^2(f_i, f_{c_k}) \right\}$  \\
  \STATE Assign $f_i$ to cluster $k_i$
 \ENDFOR\\
 \texttt{This creates a disjoint partition of the data into subsets $f_1, f_2,\ldots,f_K$.}
 \FOR {$k = 1$ to $K (=${\rm total number of clusters}) }
 \STATE Update medoid \\
 $f_{c_k} = \arg \left\{ \displaystyle \min_{f \in f_k} \sum_{f' \in f_k} W_2^2(f, f') \right\}$\\
 \ENDFOR\\
\texttt{Repeat  steps 2 to 8 using new $c_k$'s until convergence.}\\
 \STATE Group data points using the final  distribution groups in $f_1, f_2,\ldots,f_K$
 \end{algorithmic}
\end{algorithm}
\begin{algorithm}
\caption{Distributional $K$-medoids using ED (EKMd)} \label{algm4}
 %\hspace*{\algorithmicindent}
 \begin{algorithmic}[3]
\REQUIRE N distributions, $f_i\sim f_1,f_2, \ldots,f_{N}$
\STATE  Steps same as Algorithm 3, with the following, changed: steps~3~and~7:
\newline
  \hspace*{-.15in}3:$\qquad k_i=\arg \left\{\displaystyle \min_{1\leq k\leq K}  d_{f_i,f_{c_k}}^2\right\}$ 
 \newline 
\hspace*{-.15in}7:$\qquad f_{c_k} = \arg \left\{ \displaystyle \min_{f \in f_k} \sum_{f' \in f_k} d_{f, f'}^2  \right\}$ 
 \end{algorithmic}
\end{algorithm}

\subsection{\bf{Computational Complexity and Scalability}}
In general, the computational complexity of  $K$-means clustering is $\mathcal{O}(nNKT)$ and that of $K$-medoids $\mathcal O(nN^2KT)$,  where $n$ is the data dimension, $N$ is the number of data elements to be clustered, $K$ is the number of clusters, and $T$ is the number of iterations employed. ($T$ in the worst case can be exponential, leading to worst-case complexity of $n^{\mathcal O(nK)}$ \cite{vattani2009k}.) Additionally, there is $O(nN^2)$ complexity of finding pairwise distances. In the case of distributional clustering, there is the added task of estimating the distribution parameters, whose complexity is $\mathcal O(nm^2M)$, where $m$ is the number of data points per distribution, and $M$ is the number of distributions (implying a total of $N=mM$ data points). Thus the complexities of $K$-means and $K$-medoids for raw-data clustering are $\mathcal O(nmMKT)+\mathcal O(nm^2M^2)$ and $\mathcal O(nm^2M^2KT)+\mathcal O(nm^2M^2)$ respectively, and those for distributional clustering are $\mathcal O(nMKT)+\mathcal O(n^2M)+\mathcal O(nM^2)$ and $\mathcal O(nM^2KT)+\mathcal O(nm^2M)+\mathcal O(nM^2)$ respectively. These quadratic computational complexities suggest their scalability. Also, since $\mathcal O(nMKT)+\mathcal O(nm^2M)+\mathcal O(nM^2)<\mathcal O(nmMKT)+\mathcal O(nm^2M^2)$ and similarly since $\mathcal O(nM^2KT)+\mathcal O(nm^2M)+\mathcal O(nM^2)<\mathcal O(nm^2M^2KT)+\mathcal O(nm^2M^2)$, it follows that the distributional clustering has a lower time-complexity for both $K$-means and $K$-medoids compared to the raw-data clustering. Yet we show below that the accuracy of distributional clustering is higher than that of raw-data clustering.

\section{Results and Discussion}
 We implemented the distributional clustering algorithms to cluster synthetic and real-world noisy data---weather and stock data. First, the data is cleaned by removing unwanted attributes and ensuring an equal number of remaining attributes with no missing attribute values. We then extract the underlying distributions by estimating from data from the same random variable its distributions parameters---means and covariances for Gaussians (weather data) and the means of covariances of the natural logarithms for the lognormals (stock data). 
For the case of the real-world weather data, we treat each season of each year to be a Gaussian distribution, and accordingly, we have $4$ distributions per year. For the case of the real-life stocks data, we model each stock to be a lognormal distribution, considering $77$ total stocks picked from the Nasdaq top-100 for the years 2018-19. The performance of the six different clustering algorithms---the classical versions of $K$-means and $K$-medoids and their $W_2$ and ED-based extensions, namely, WKM, WKMd, EKM, EKMd---are compared using the measures of Accuracy, NMI, and ARI as described next.

\subsection{Performance metrics}
The accuracy of the six clustering techniques: classical $K$-means (KM), $W_2$ $K$-means (WKM), ED $K$-means (EKM), classical $K$-medoids (KMd),  $W_2$ $K$-medoids (WKMd), and ED $K$-medoids (EKMd), are compared using the following defined three commonly used performance metrics of Accuracy, NMI (normalized mutual information), and ARI (adjusted rand index). They all assume the existence of the ground truth clustering, denoted $C^*$, to compare against the computed clustering, $C$, and compute a normalized score within the unit interval, with 1 being the maximum accuracy score. Given $S$, a set of $N$ data points, and its two $K$-sized cluster partitions, the computed one $C$ and the ground truth $C^*$:
\[ C = \{S_C(1),\ldots,S_C(K)\};\; C^* = \{S_{C^*}(1),\ldots,S_{C^*}(K^*)\},\]
define  $n_{ij} := |S_C(i) \cap S_{C^*}(j) |,  \quad a_i := \sum_{j=1}^{K^*} n_{ij}, \quad  b_j = \sum_{i=1}^K n_{ij}$.
Note $n_{ij}$ denotes the number of data points common between the clusters $X_{C}(i)$ and $X_{C^*}(j)$. 
\begin{enumerate}
\item \textbf{accuracy} is simply the ratio of the correctly clustered data points to the total number of data points:
\begin{align*}\textbf{Accuracy}  = %\max_{perm\in P}
\frac{1}{N}\sum_{i=1}^{\min\{K,K^*\}} n_{ii}.\end{align*}
       
\item \textbf{Normalized Mutual Information (NMI) } \cite{danon2005comparing}
The mutual information $I(C^*; C)$ between the two clusterings is used to compute the two normalized indices $\frac{I(C^*; C)}{H(C^*)}$ and  $\frac{I(C^*; C)}{H(C)}$, respectively, whose harmonic mean gives the desired index:
\begin{align*}
      &\textbf{NMI}=2\frac{I(C^*; C)}{H(C^*)+H(C)};\\
      &I(C^*;C) = H(C^*)-H(C^*|C)\\
      &= -\sum_{c^* \in C^*} p_{C^*}(c^*) \log_2  p_{C^*}(c^*) - \sum_{c \in C}p_{C}(c)H(C^*|C=c)\\
     &=  -\sum_{j=1}^{K^*} \frac{b_j}{n}\log_2\frac{b_j}{n}-  \sum_{i=1}^K \frac{a_i}{n} \sum_{j=1}^{K^*} \frac{n_{ij}}{a_i} \log_2  \frac{n_{ij}}{a_i}.
\end{align*}
     
\item \textbf{Adjusted Rand Index (ARI) } \cite{hubert1985comparing}
The  Rand Index (RI) computes a similarity measure between two clustering by counting samples in all pairs of cells taken from the two clusters. The ARI score is then the adjusted version of RI, "corrected-for-chance," and normalized:    
\begin{align*}   
&\textbf{ARI} = \frac{\rm{RI }   -\rm{Expected(RI) }}{\rm{Max(RI)} - \rm{Expected(RI) }}\\ 
&=   \frac{\sum_{ij} \binom{n_{ij}}{2} - 
   \left( \frac{ \sum\limits_{i=1}^K \binom{a_i}{2} \sum\limits_{j}^{K^*} \binom{b_j}{2}}{ \binom{n}{2}} \right)}
   {\frac{1}{2} \left( \sum\limits_{i}^K\binom{a_i}{2}+ \sum\limits_{j}^{K^*}\binom{b_j}{2} \right)
   - \left( \frac{\sum\limits_{i}^K\binom{a_i}{2} \sum\limits_{j}^{K^*}\binom{b_j}{2}}{\binom{n}{2}} \right)}.
 \end{align*}
\end{enumerate}

\subsection{Synthetic data with unbalanced clusters}
\begin{figure}[!htb]
    \centering
    \begin{subfigure}[t]{0.15\textwidth}
        \centering
        \includegraphics[height=0.9in]{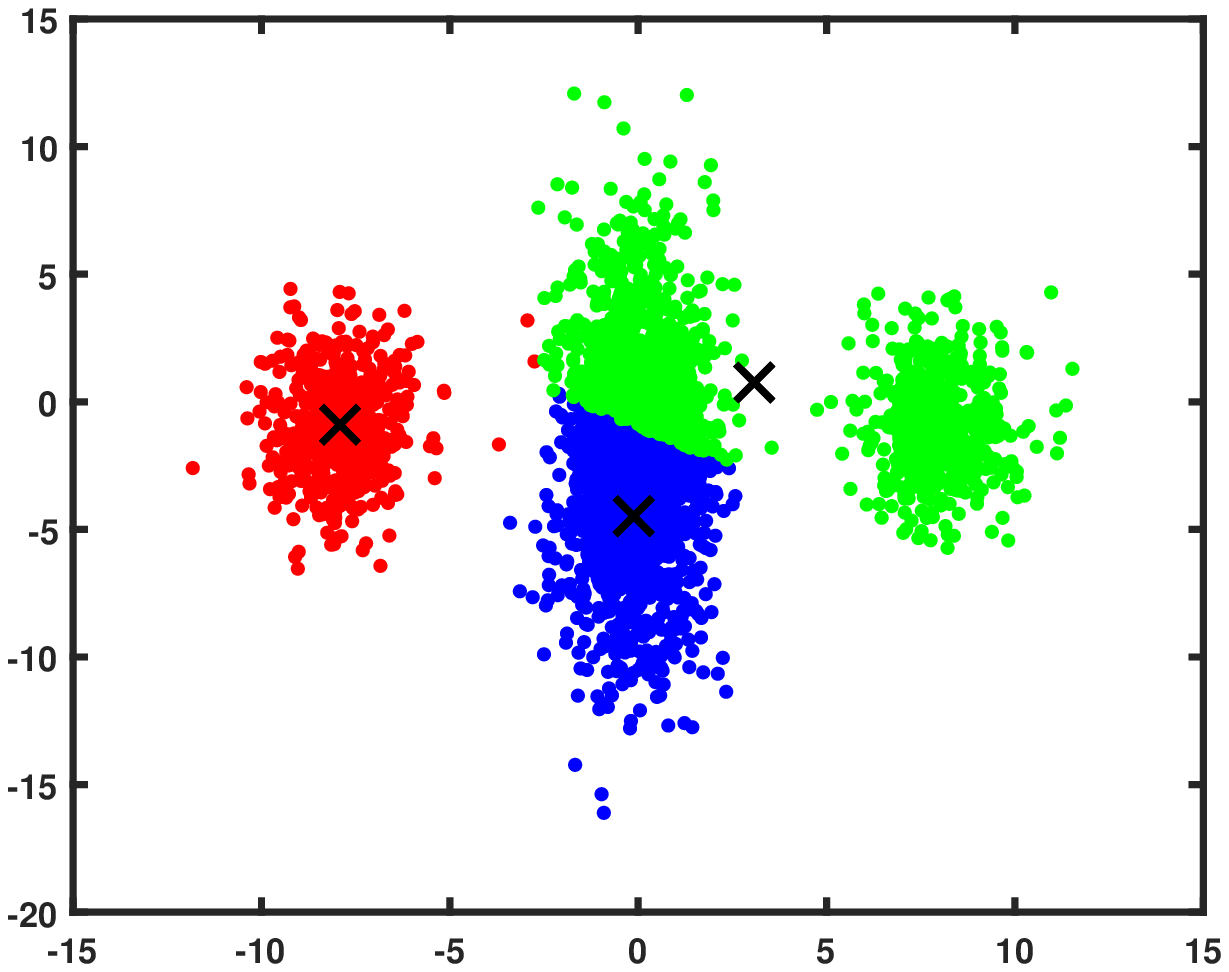}
         \caption{}
    \end{subfigure}%
    ~ 
    \begin{subfigure}[t]{0.15\textwidth}
        \centering
        \includegraphics[height=0.9in]{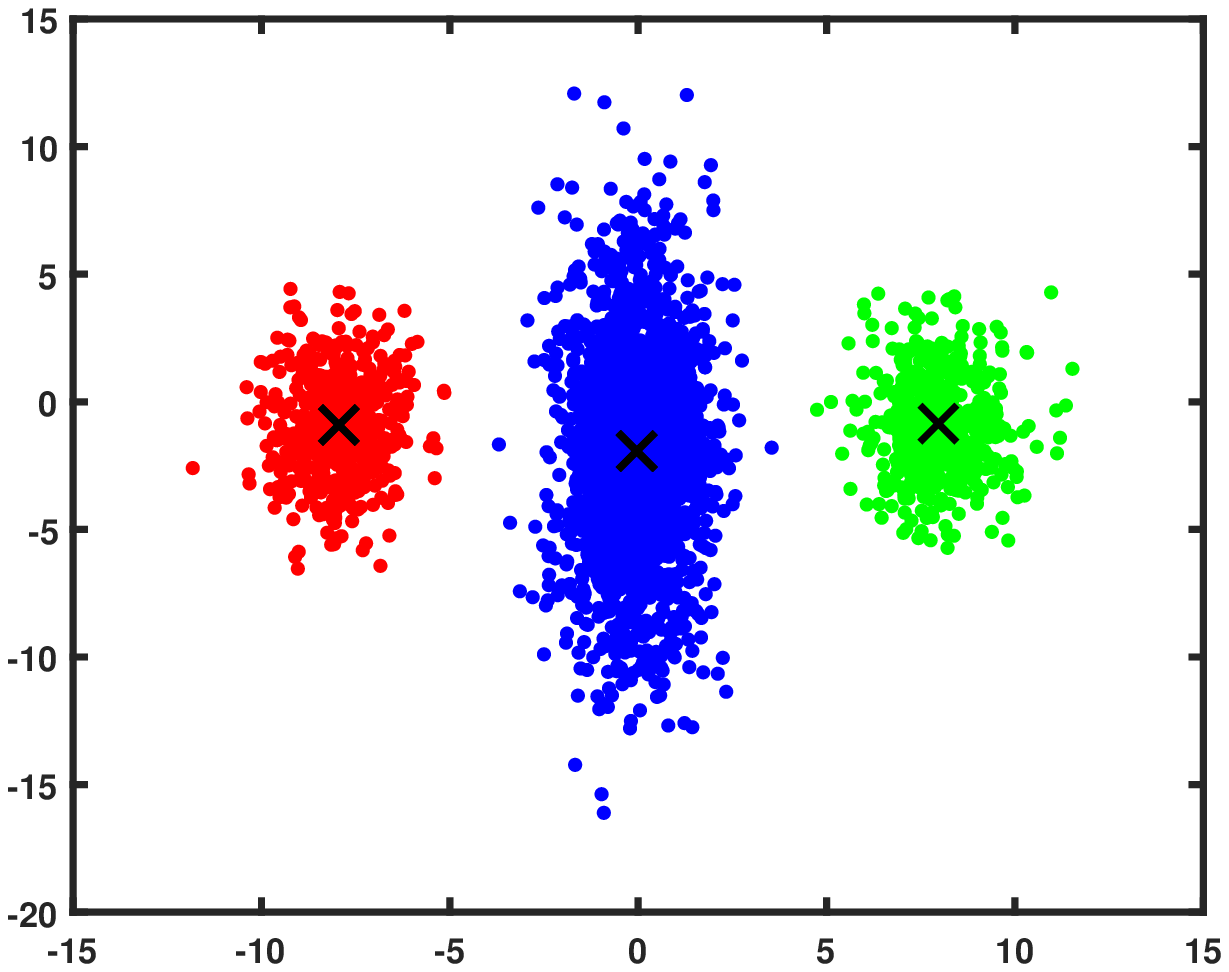}
         \caption{}
    \end{subfigure}
    ~ 
     \begin{subfigure}[t]{0.15\textwidth}
        \centering
        \includegraphics[height=0.9in]{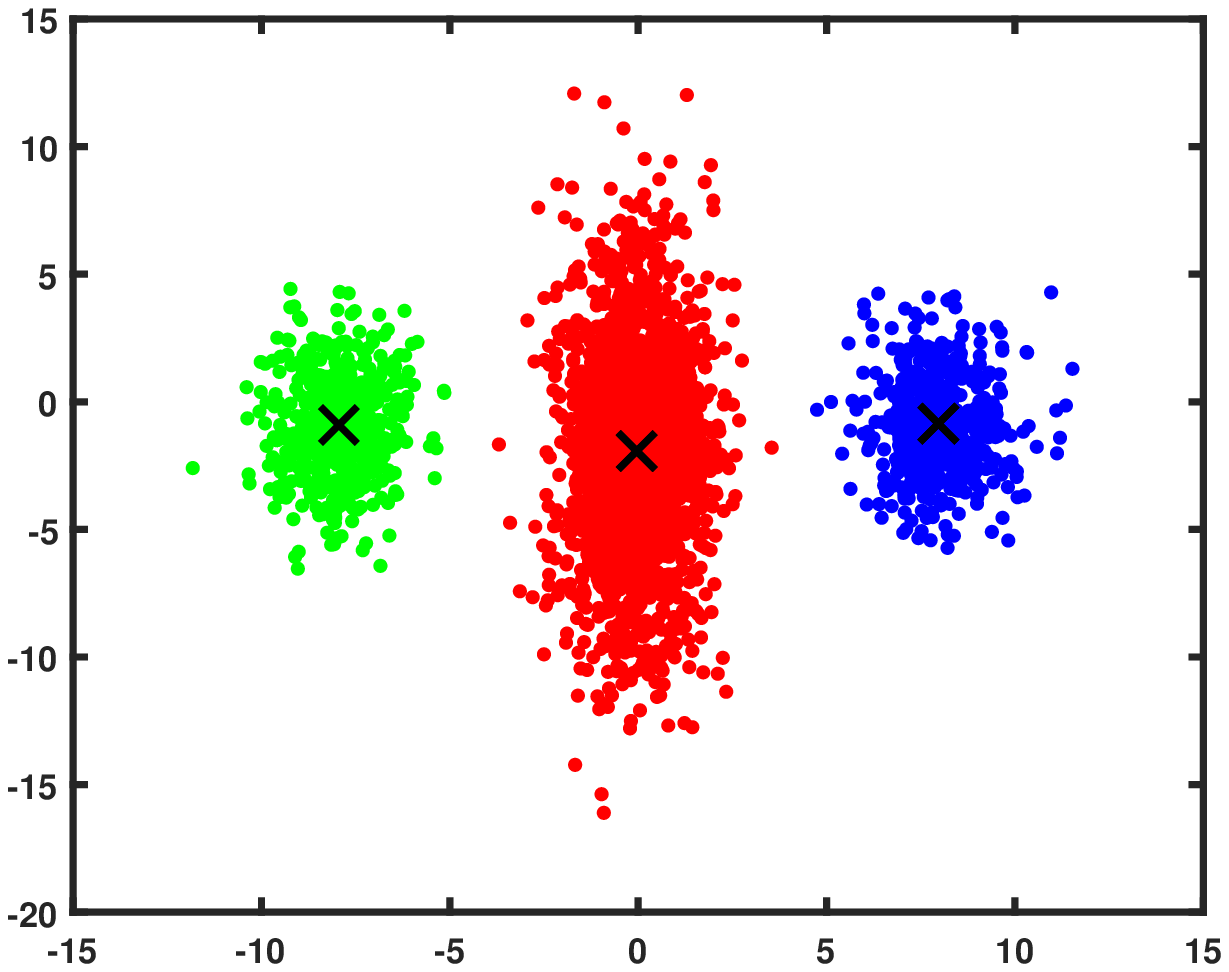}
         \caption{}
    \end{subfigure}
   ~
    \begin{subfigure}[t]{0.15\textwidth}
        \centering
        \includegraphics[height=0.9in]{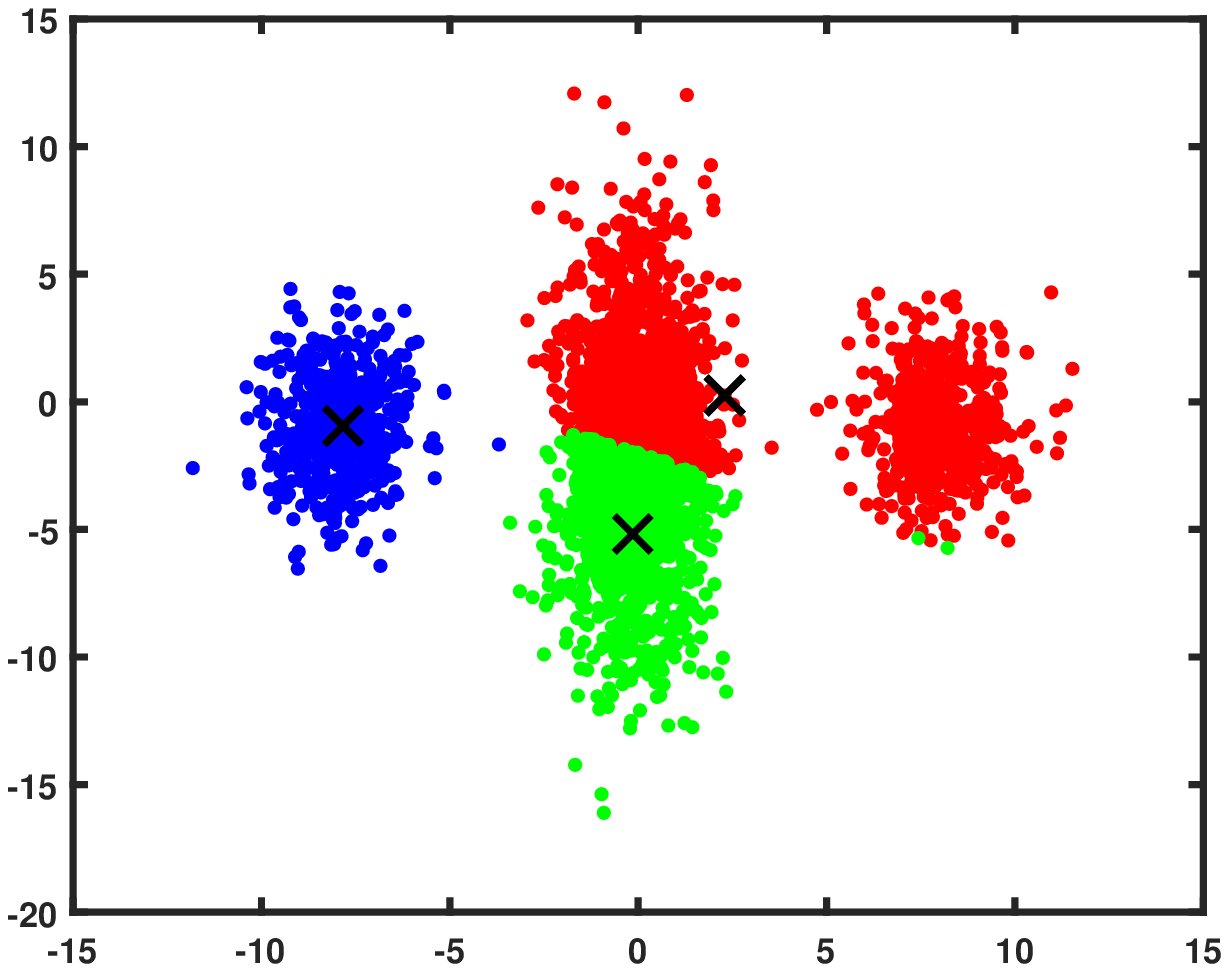}
         \caption{}
    \end{subfigure}%
    ~ 
    \begin{subfigure}[t]{0.15\textwidth}
        \centering
        \includegraphics[height=0.9in]{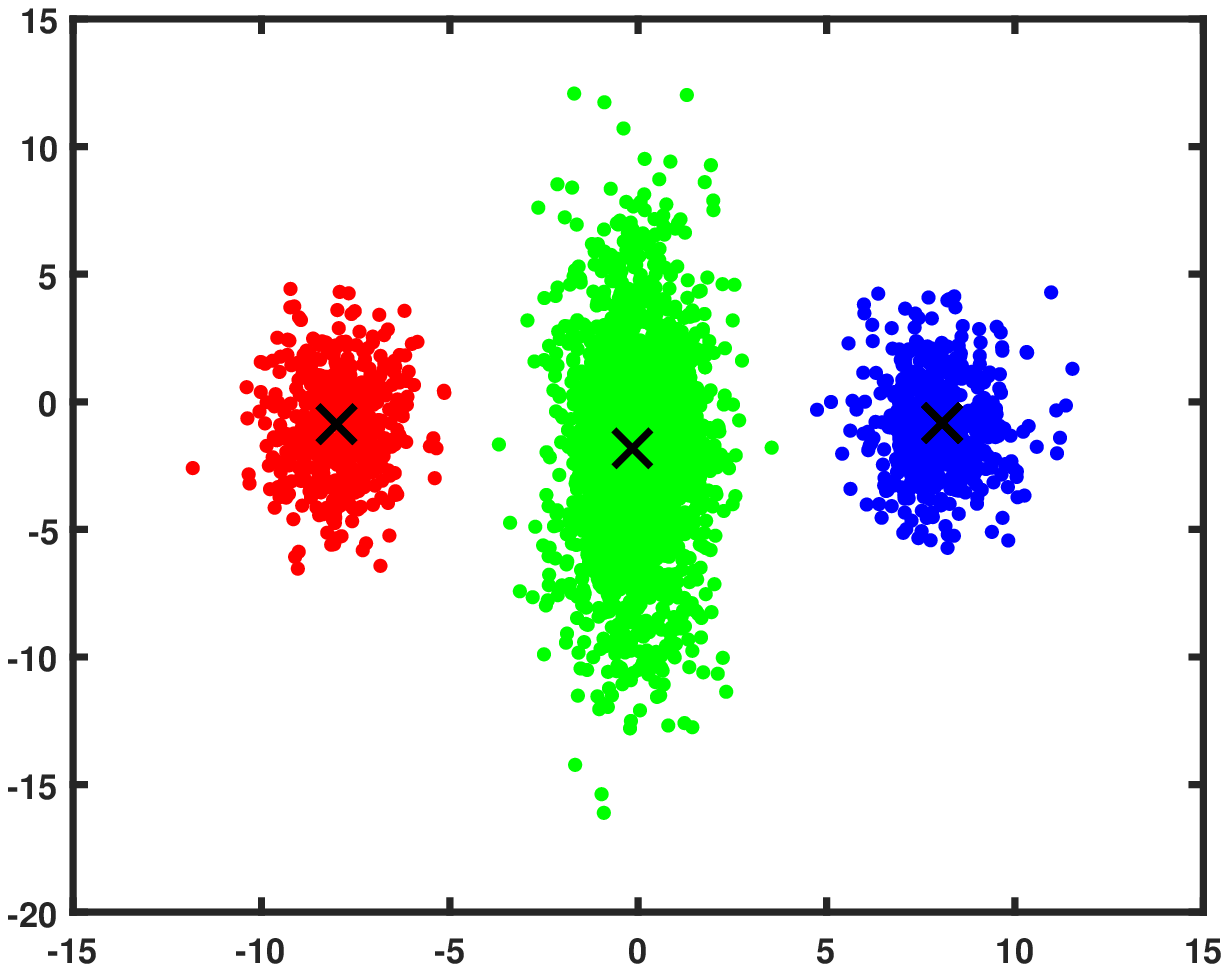}
         \caption{}
    \end{subfigure}
    ~ 
     \begin{subfigure}[t]{0.15\textwidth}
        \centering
        \includegraphics[height=0.9in]{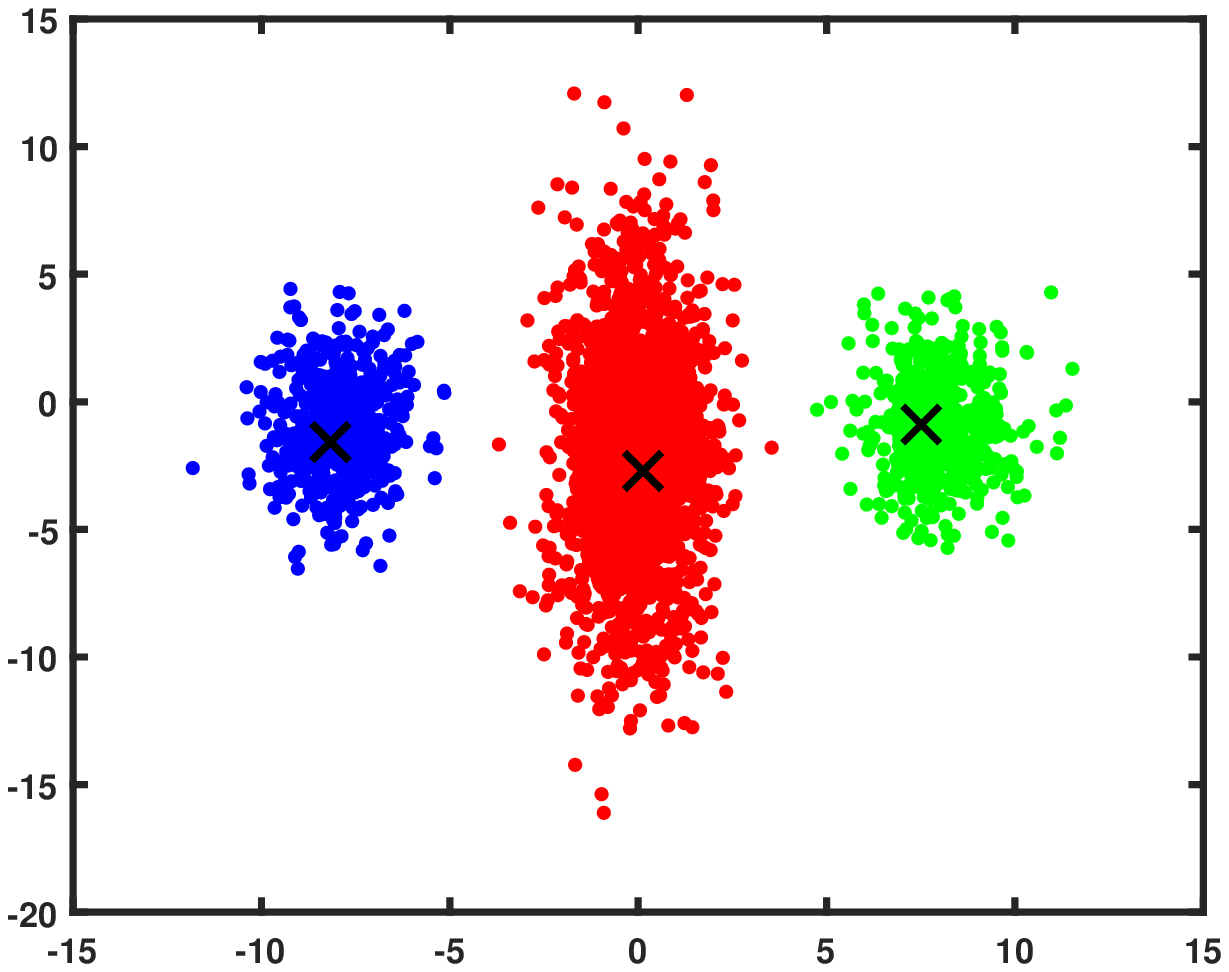}
         \caption{}
    \end{subfigure}
    \caption{Synthetic data result: Plots of data clusters (dots) and cluster-centers (black crosses)  using  (a) KM, (b) WKM, (c) EKM, (d) KMd, (e) WKMd, (f) EKMd.}
    \label{syn}
\end{figure}

Clusters are termed unbalanced if their sizes are disparate. It is known that classical clustering does not work well with imbalanced-sized data  \cite{6121900}.
To illustrate that our algorithm is robust enough to cluster even imbalanced-sized data, a synthetic dataset is generated using random Gaussian distribution, \textit{randn}, in $2$-dimensional space, having $3000$ samples, grouped into 3 groups of $2000 + 500 + 500$:
 \begin{align*}
  X \sim   \begin{bmatrix} randn(2000,1)& 4\times  randn(2000,1)-2\\ randn(500,1)-8 & 2\times randn(500,1)-1\\ randn(500,1)+8&2\times randn(500,1)-1
  \end{bmatrix}.
\end{align*} 
The raw-data, as well as the inferred distributions, needs to be grouped into $3$ clusters per the $3$ distributions shown above. The marginal and joint distributions were computed over $20$ samples belonging to the same clusters; hence, the data-distribution has $2000/20+500/20+500/20=100+25+25=150$ entries. The results obtained from the six clustering algorithms are visually shown in Fig.~\ref{syn} and evaluated in Table.~\ref{tab:syn}. Even though the three clusters are visibly obvious, the classical KM and KMd failed to produce accurate clusters; in contrast, the accuracy of the other 4 methods is $100\%$, exhibiting the robustness of the distribution-data-based clustering for unbalanced clusters. 

In Table~\ref{tab:syn}, we also see that WKM and EKM methods have the same clusters and hence also the same means and variances of their cluster-centers, as explained theoretically. For the cases of WKMd and EKMd, the cluster-center indices differ from each other, as also expected, even though the clusters are the same, owing to the dependence of the cluster-centers on the distance measures employed.
\setlength{\textfloatsep}{6pt}
  \begin{table}[!htb]
     \centering
      \resizebox{\columnwidth}{!}{
    \begin{tabular}{|c|c|c|c|c|}
\hline
 & Accuracy & NMI & ARI& Centers \\
\hline
KM & $0.7262$ & $0.6036$&$0.6932$&\\
\hline
WKM& $1.0000$&$1.0000$&$1.0000$&
$\begin{bmatrix}-0.0391&-1.9316 \end{bmatrix}$\\
&&&&$\begin{bmatrix} -7.9458& -0.9100 \end{bmatrix}$\\
&&&&$\begin{bmatrix} 7.9681 & -0.8576 \end{bmatrix}$\\
\hline
EKM&$1.0000$&$1.0000$&$1.0000$& $\begin{bmatrix}-0.0391&-1.9316 \end{bmatrix}$\\
&&&&$\begin{bmatrix} -7.9458& -0.9100 \end{bmatrix}$\\
&&&&$\begin{bmatrix} 7.9681 & -0.8576 \end{bmatrix}$\\
\hline
KMd& $0.6543$& $0.5759$ & $0.6625$ &\\
\hline
WKMd&$1.0000$&$1.0000$&$1.0000$& $(69, 144, 119)$\\
\hline
EKMd &$1.0000$&$1.0000$&$1.0000$& $(132, 117, 55)$\\
\hline
    \end{tabular} }
    \caption{Table of Evaluation measures for synthetic data}
    \label{tab:syn}
     \end{table}
Table~\ref{tab:syn} shows that the distribution-based clusterings outperform the classical ones. As can be seen, there is a dramatic improvement over the classical clustering ($0.7262$ for KM and $0.6543$ for KMd, whereas $1$ for WKM, EKM, WKMd, and EKMd) even for the case of imbalanced-sized data.

\subsection{Real-world weather data}
To demonstrate the performance of the explained clustering algorithms and show that the distribution-based algorithms work better in the case of uncertain data, we applied them to real-life weather data sourced from \href{https://coagmet.colostate.edu/rawdata_form.php 
}{\textit{ Colorado State University CoAgMET Raw Data Access}}, that were collected from five weather stations listed in Table~\ref{WD}.  
  \begin{table}[!htb]
    \centering
     \resizebox{\columnwidth}{!}{
    \begin{tabular}{|c|c|c|r|}
\hline
&Station& Number of years obtained& Data Length\\
\hline
A&Avondale& $22$ & $7392$\\
\hline
B&Ault& $19$ & $6384$ \\
\hline
C&Dove Creek&$21$&$7056$\\
\hline
D&Fort Collins &$23$ &$7728$\\
\hline
E&Kirk &$20$ &$6720$\\
\hline
 \multicolumn{2}{|c|}{ Total} &$105$ &$35280$\\
\hline
    \end{tabular}} \caption{Weather Data details}
    \label{WD}
    \end{table}
    \setlength{\textfloatsep}{6pt}
For an even computation of data-distributions, we kept only $28$ entries from each month, and accordingly, the data lengths, based on the number of years, are as listed in Table~\ref{WD}. 

Data in the same meteorological seasons in the USA were considered to be in the same cluster: Spring: 03/01 -  05-31;	 Summer: 06/01 - 08/31; Fall: 09/01 - 11/30; Winter: 12/01 - 02/28. Thus, we have $4$ clusters in total, and the ground truth cluster for each measurement was acquired based on the date and the corresponding season for each entry. For each weather station, the data for the same year's season, with each season containing 84~days (28~days/month for 3 months in a season) worth of data, was treated as a random variable, thereby producing $4\times $ \#Station $\times$ \#Years $=4\times 5\times 21=420$ of total random variables to be clustered into 4 seasons, with each random variable supported by 84 days of data. Thus there were a total of $420\times 84=35820$ data points. It was reasonably assumed that the weather within a season of a year at a location follows Gaussian distribution.

\begin{figure}[!htb]
    \centering
    \begin{subfigure}[t]{0.1\textwidth}
        \centering
        \includegraphics[height=0.6in]{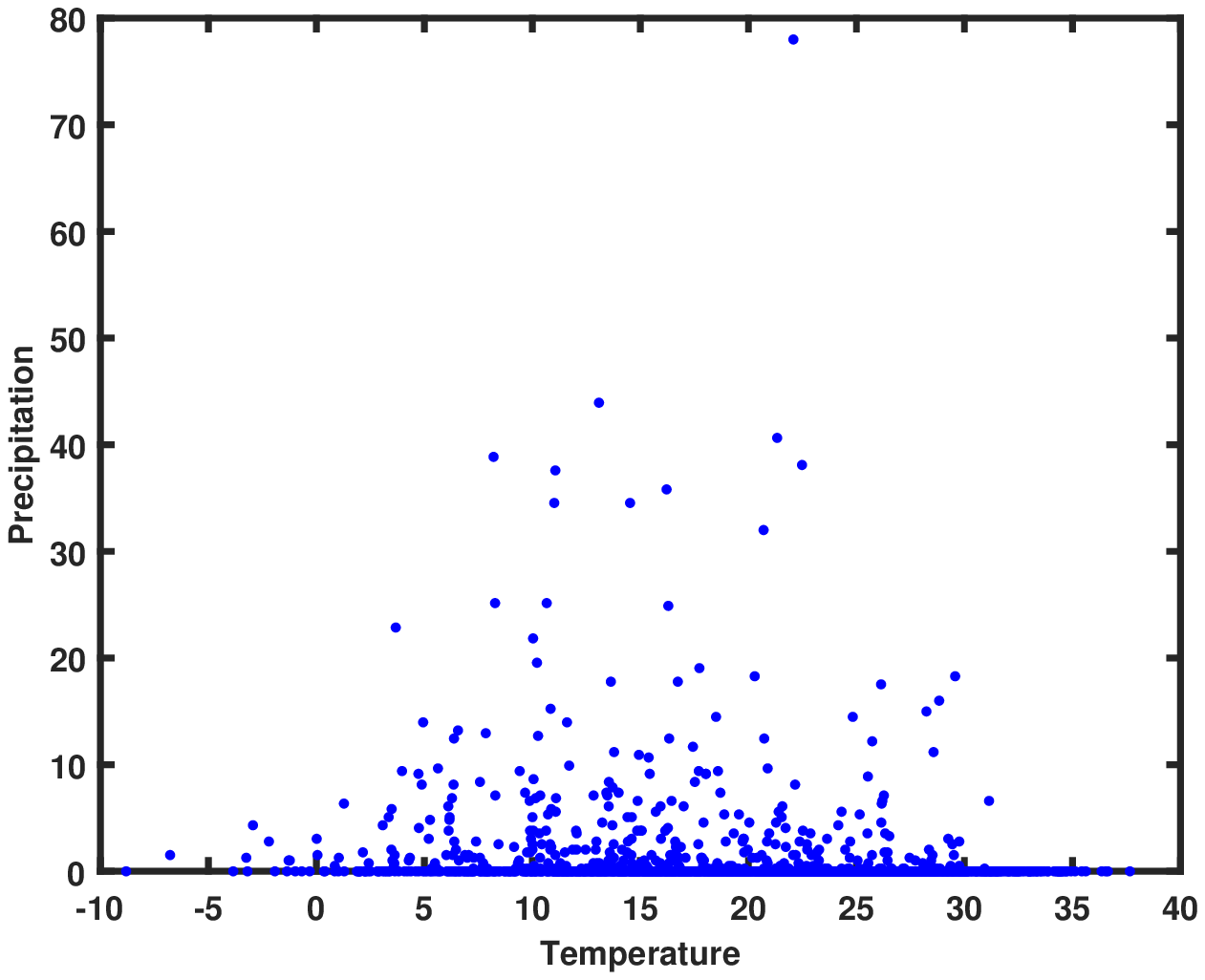}
        \subcaption{Spring}
    \end{subfigure}%
    ~ 
    \begin{subfigure}[t]{0.1\textwidth}
        \centering
        \includegraphics[height=0.6in]{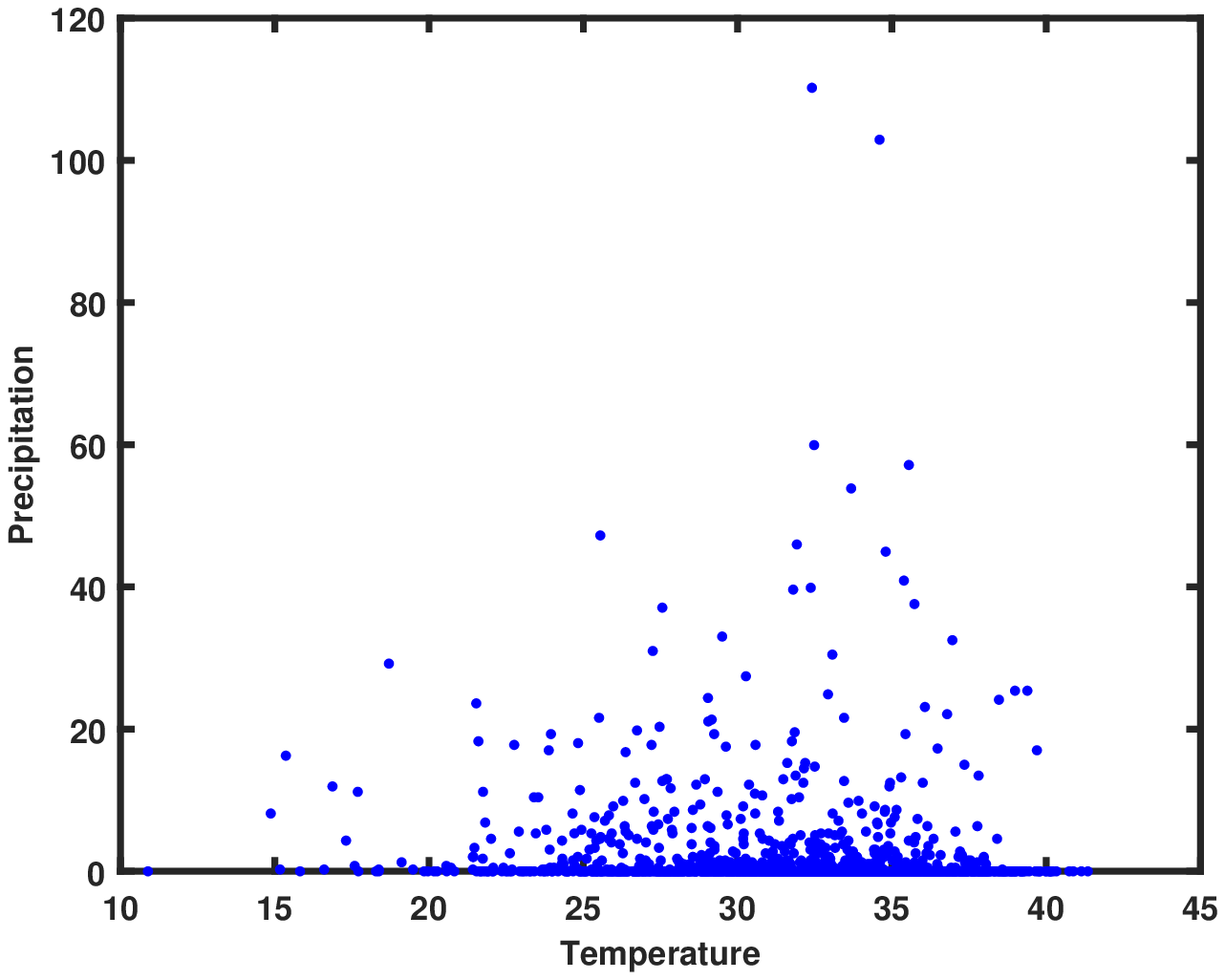}
        \subcaption{Summer}
    \end{subfigure}
    ~ 
     \begin{subfigure}[t]{0.1\textwidth}
        \centering
        \includegraphics[height=0.6in]{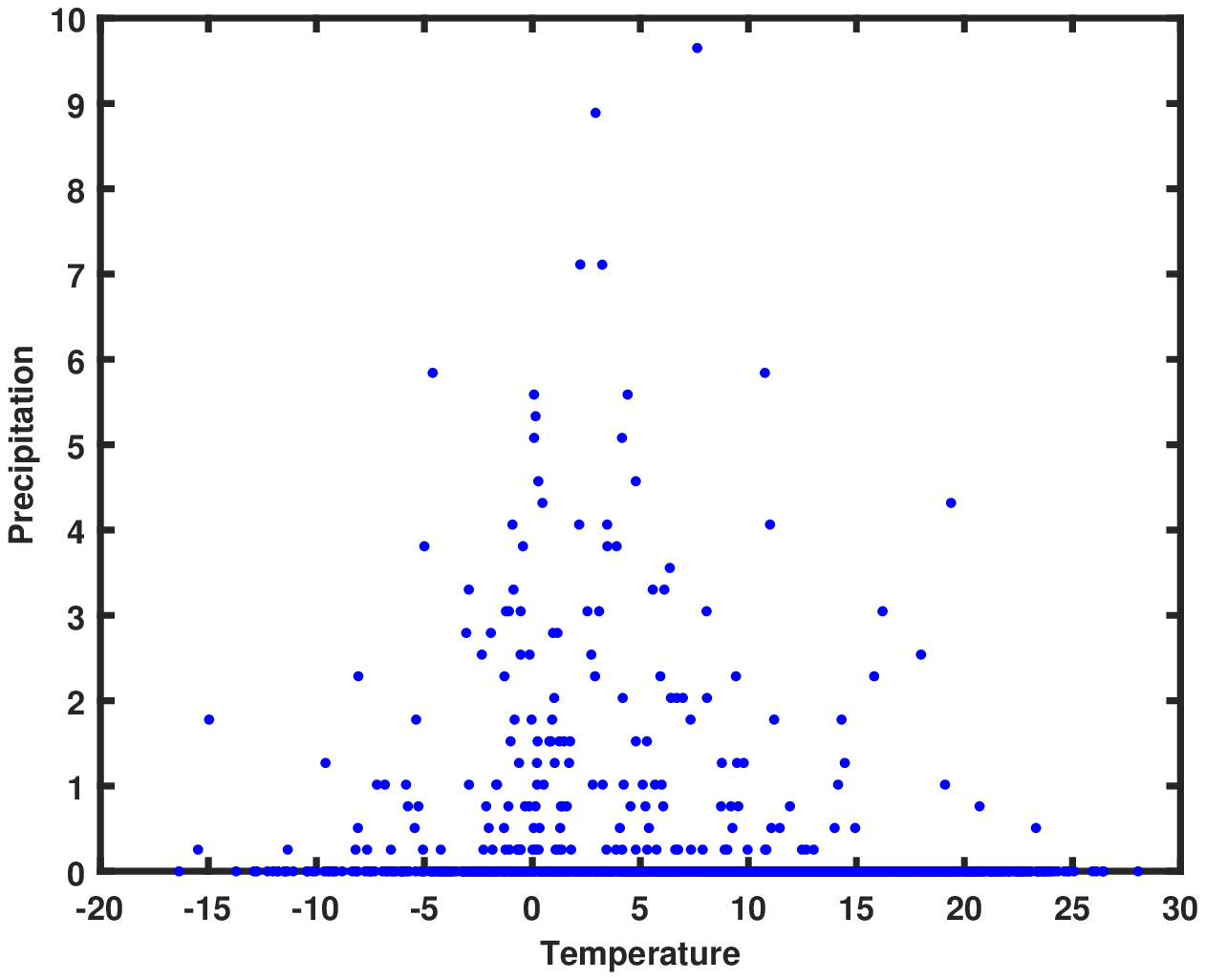}
        \subcaption{Winter}
    \end{subfigure} 
    ~ 
     \begin{subfigure}[t]{0.1\textwidth}
        \centering
        \includegraphics[height=0.6in]{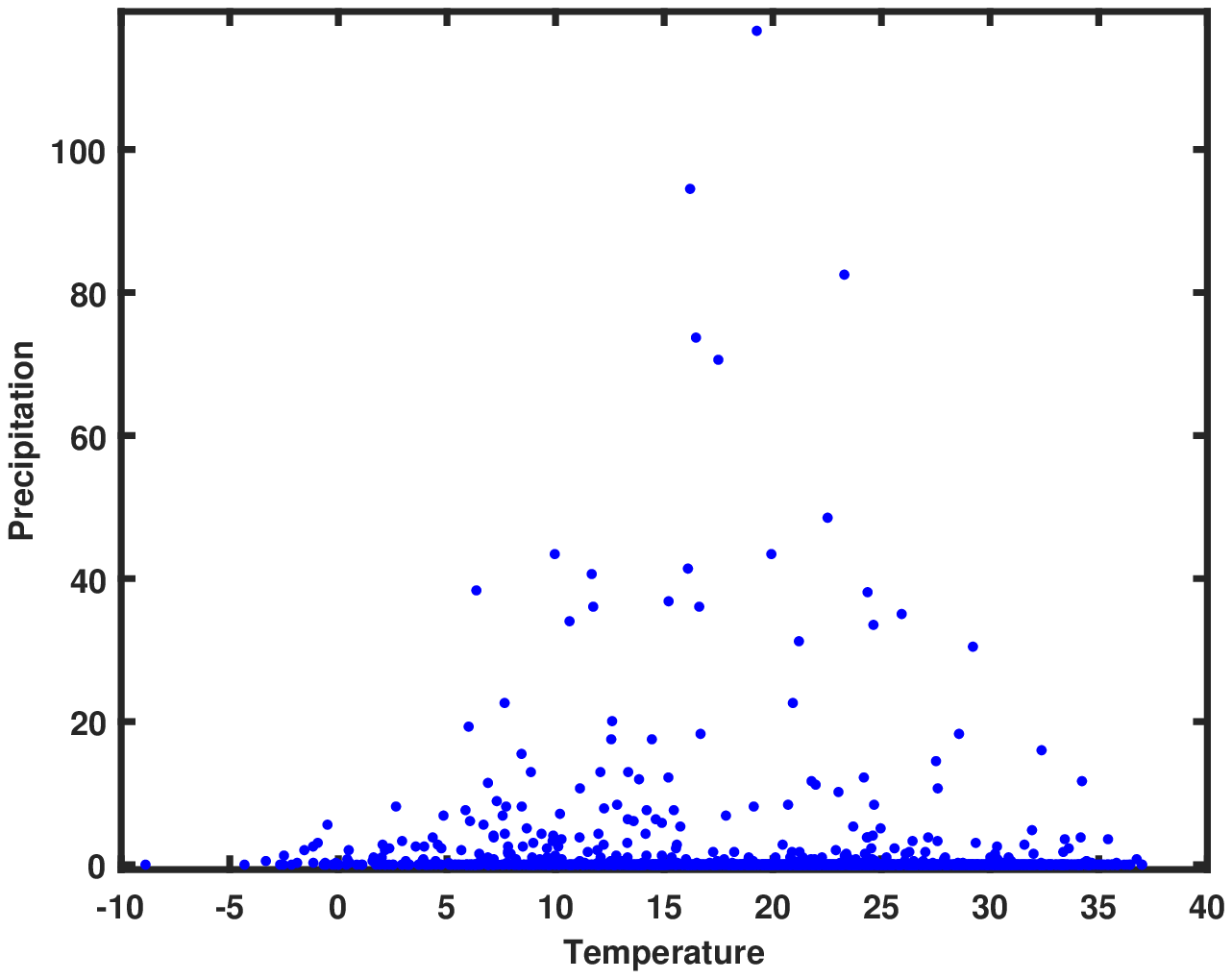}
        \subcaption{Fall}
    \end{subfigure}
    \caption{ Data-distribution of each of the four seasons for the Avondale station}
    \label{Av:szn}
    \end{figure}

Fig.~\ref{Av:szn} shows the data-distribution of each of the four seasons for the Avondale station, where for easier visualization, we only plot the $2$-dimensional temperature and precipitation data for all the weather data plots. It can be easily seen that the data contains outliers.

\subsubsection{\texorpdfstring{3}{}-Dimensional weather data analysis}

In this part of the study, we analyzed $3$-dimensional weather data consisting of three features: {\em maximum daily temperature ($^\circ$C), precipitation (mm), and vapor pressure (kPa)}, extracted from the weather data from five stations: Avondale, Ault, Dove Creek, Fort Collins, and Kirk, between 1992 and 2021. As noted above there are 420 random variables, with each random variable having 84 days/season of 3D-data, implying a data set of $35,280 \times 3$ entries. In the preprocessing stage, we computed $420$ numbers of 3D means and $3\times 3$ variances, and $420^2$ $3\times 3$ covariances, which took $1.9194$ sec.

\begin{figure}[!htb]
    \centering
    \begin{subfigure}[t]{0.15\textwidth}
        \centering
        \includegraphics[height=0.9in]{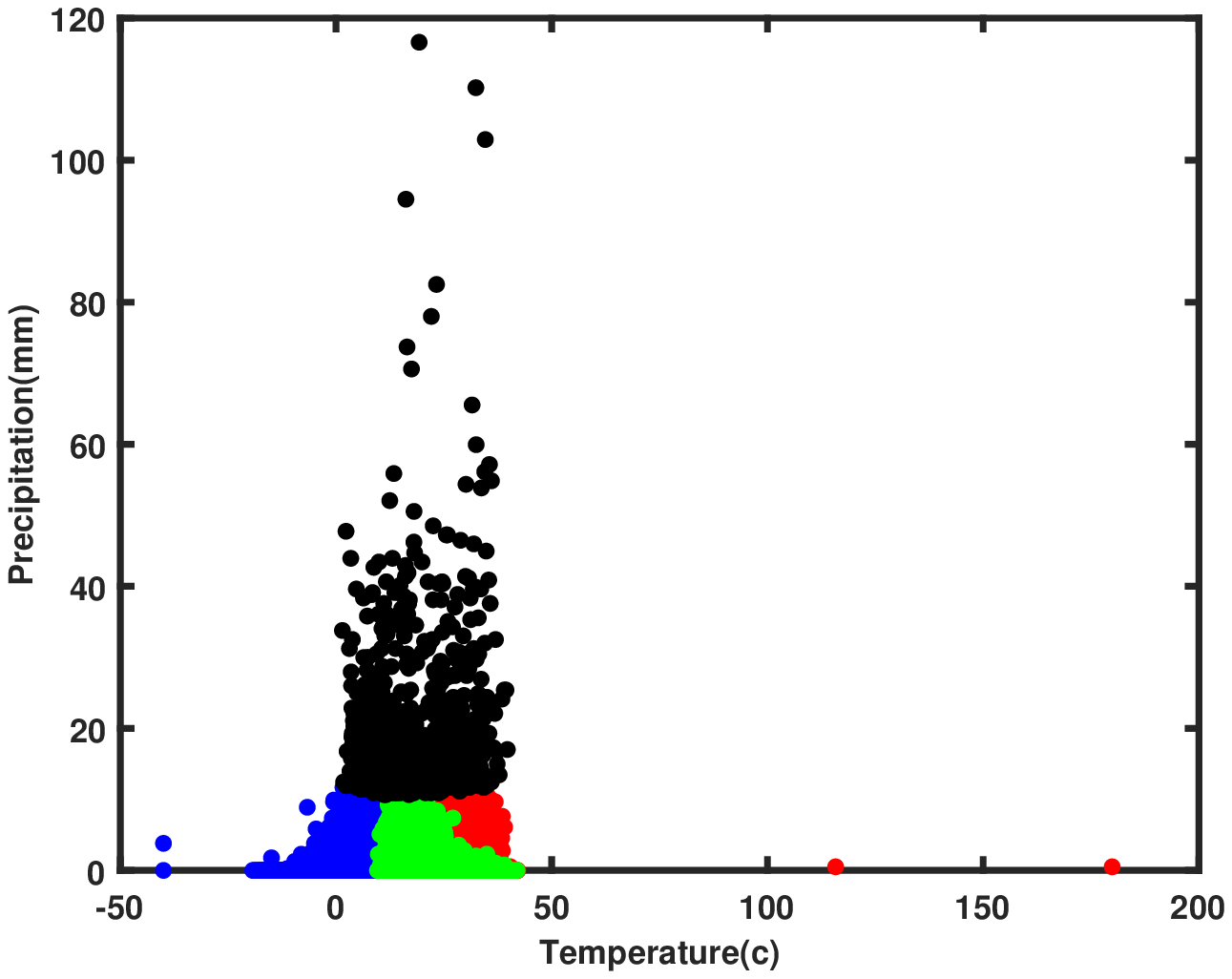}
         \subcaption{}
    \end{subfigure}%
    ~ 
    \begin{subfigure}[t]{0.15\textwidth}
        \centering
        \includegraphics[height=0.9in]{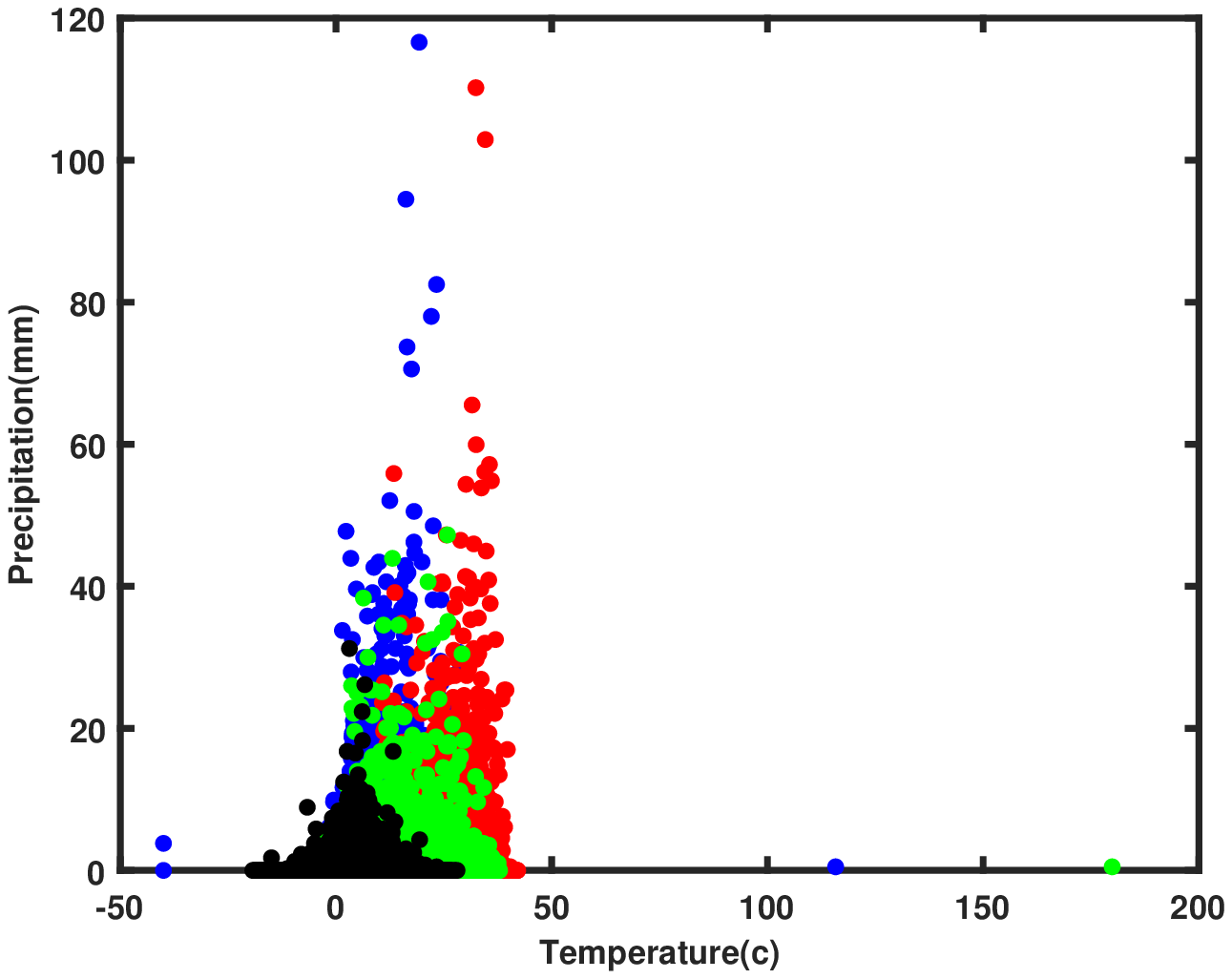}
         \subcaption{}
    \end{subfigure}
    ~ 
     \begin{subfigure}[t]{0.15\textwidth}
        \centering
        \includegraphics[height=0.9in]{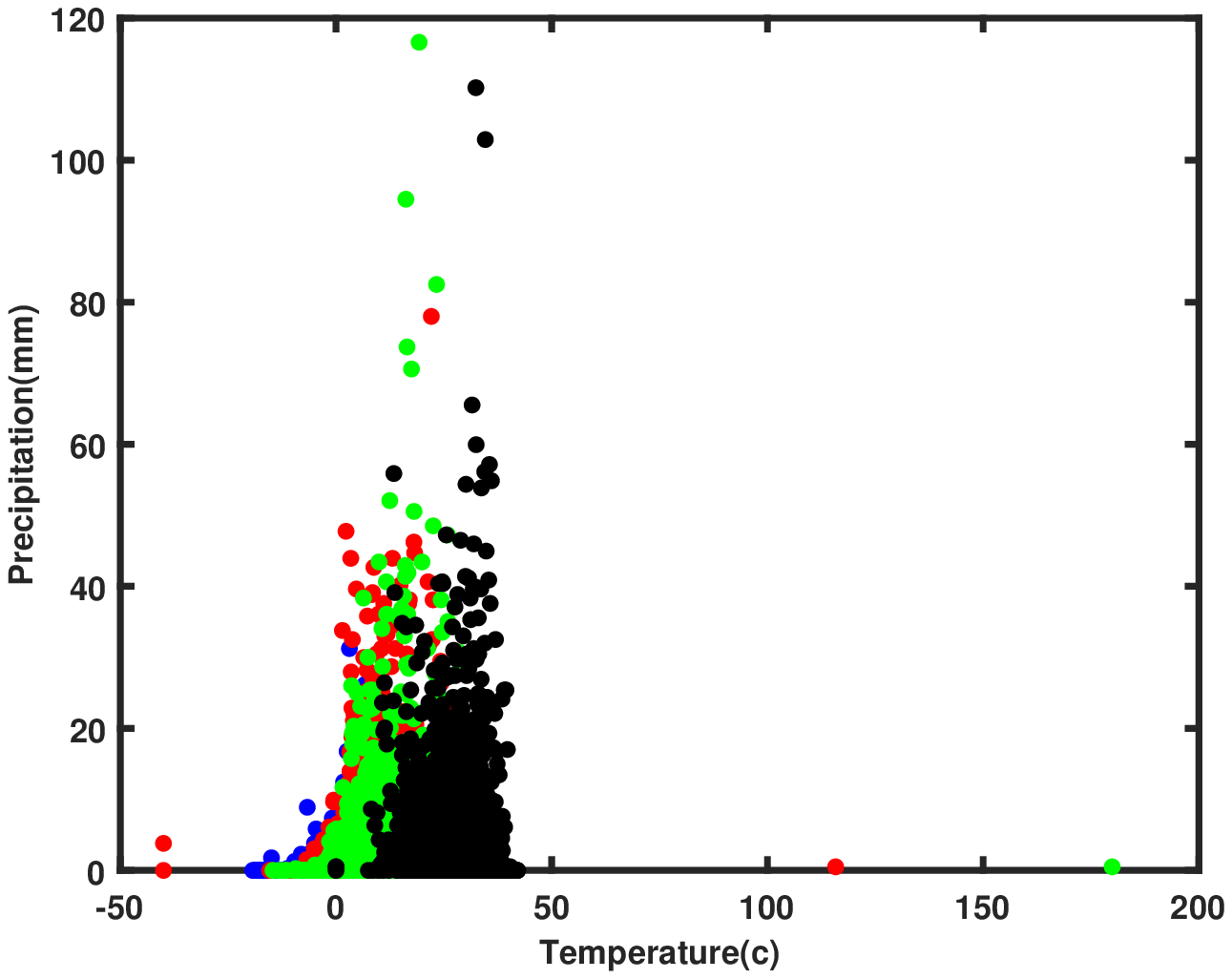}
         \subcaption{}
    \end{subfigure}
   ~
    \begin{subfigure}[t]{0.15\textwidth}
        \centering
        \includegraphics[height=0.9in]{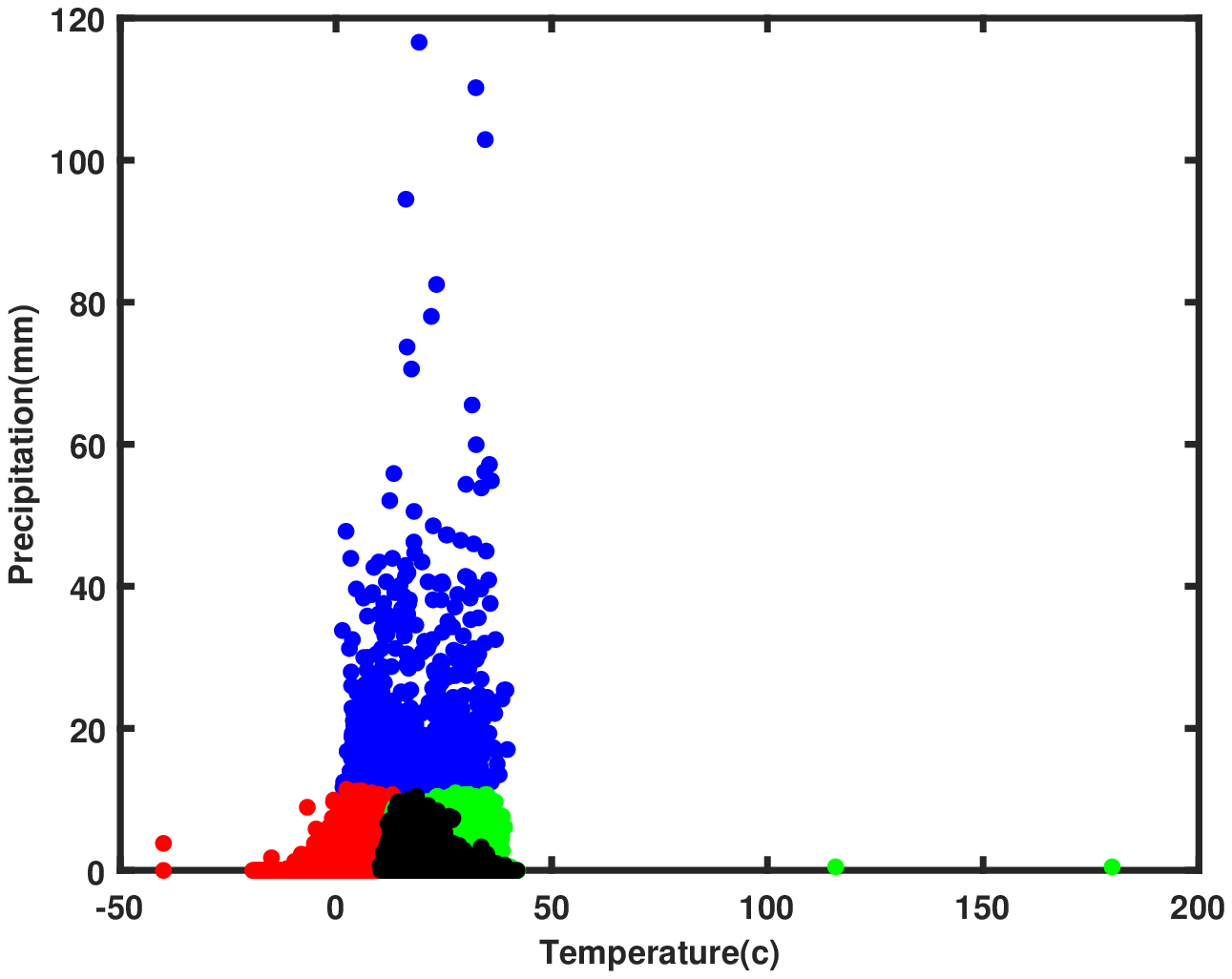}
         \subcaption{}
    \end{subfigure}%
    ~ 
    \begin{subfigure}[t]{0.15\textwidth}
        \centering
        \includegraphics[height=0.9in]{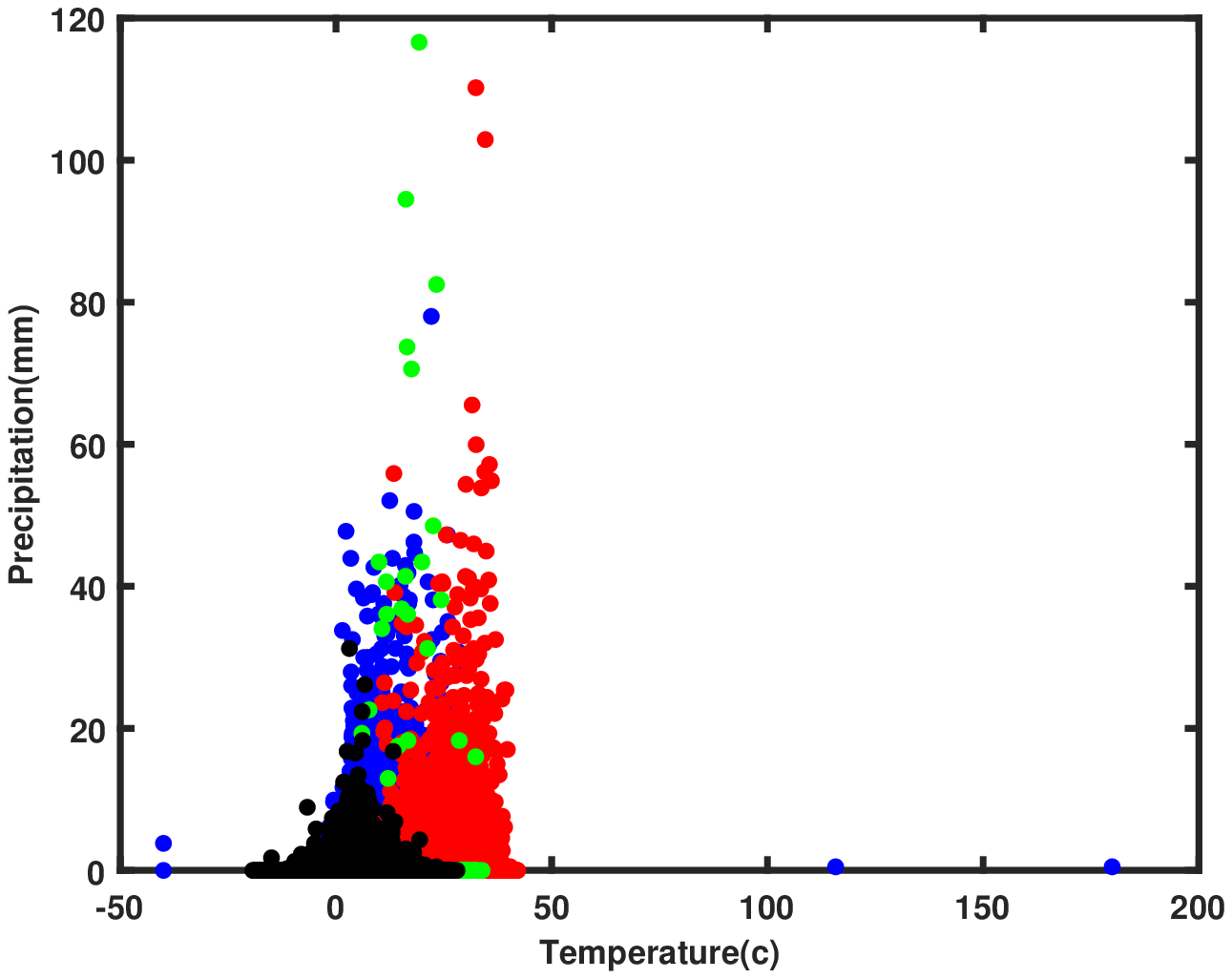}
         \subcaption{}
    \end{subfigure}
    ~ 
     \begin{subfigure}[t]{0.15\textwidth}
        \centering
        \includegraphics[height=0.9in]{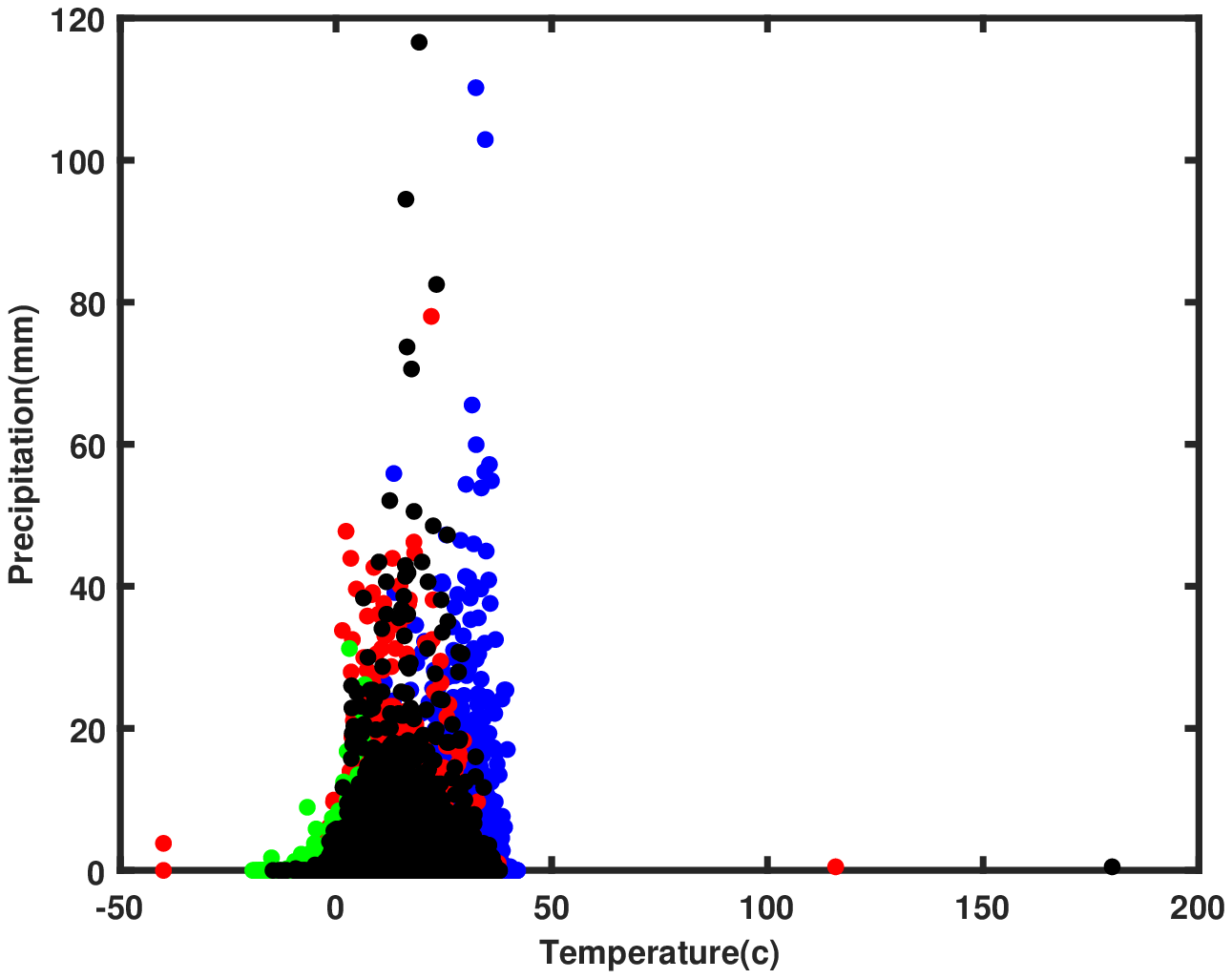}
         \subcaption{}
    \end{subfigure}
    \caption{3-D Weather data result: Plots of data clusters (dots)  using  (a) KM, (b) WKM, (c) EKM, (d) KMd, (e) WKMd, (f) EKMd.}
    \label{Cb}
\end{figure}

We used distributional clustering algorithms to analyze the data, and the performances and compute times of the clustering algorithms are shown in Fig.\ref{Cb} and Table\ref{tab:Cb}. The accuracies of KM, WKM, and EKM were $0.5649$, $0.8429$, and $1.0000$ respectively, whereas the accuracies of the corresponding $K$-medoids versions were $0.5637$, $0.8500$, and $0.9976$ respectively. The NMIs were: $0.3148$, $0.7755$, $1.0000$ and $0.3141$, $0.7799$, $0.9903$, whereas the ARIs were: $0.7052$, $0.8922$, $1.0000$ and $0.7047$, $0.8950$, $0.9976$.

The results show that the distribution-based $K$-means and $K$-medoids algorithms outperform the corresponding classical versions. The ED-based distance measurement offers higher accuracy, NMI, and ARI values over the $W_2$ based ones, which in turn performs better than the classical ones. These indicate that the algorithms are more robust to real-life noisy weather data, and the overall {\em computational time of distributional clustering is less compared to the classical ones that operate on raw-data, yet the accuracy of our algorithms remains higher}.

In summary, this study provides an efficient and accurate approach to analyzing small dimensional weather data, which can be useful in various weather-related applications.

\setlength{\textfloatsep}{6pt}
\begin{table}[!htb]
     \centering
      \resizebox{\columnwidth}{!}{
    \begin{tabular}{|c|c|c|c|c|}
\hline
 & Accuracy & NMI & ARI& Compute Time (s) \\
\hline
KM &$ 0.5649$ &$ 0.3148$ & $0.7052$   &$8.3068$\\
\hline
WKM&$0.8429$ &$ 0.7755$ & $0.8922$ &$7.6164$\\
\hline
EKM&$1.0000$&$1.0000$&$1.0000$& 6.9747\\
\hline
KMd&$ 0.5637$ & $0.3141$ & $0.7047$ &$5.3981$\\
\hline
WKMd&$0.8500$ & $0.7799$ & $0.8950$ &$9.9327$\\
\hline
EKMd &$0.9976$ & $0.9903$ & $0.9976 $&$5.0629$\\
\hline
    \end{tabular} }
    \caption{Evaluation measures for 3-D weather data}
    \label{tab:Cb}
  \end{table}

\subsubsection{Clustering of 7-Dimensional Weather Data and Comparison of Computation Times}

To compare the time complexity as the data size grows, we increased the measurement parameters of daily weather data from 3 to $7$, to include the following features: {\em mean, maximum, and minimum temperature ($^\circ$ C), vapor pressure (kPa), maximum and minimum relative humidity (Fraction), and precipitation (mm)}. The resulting data entries is larger compared to our previous experiment, having increased $7$-dimensions for the means, variances and covariances. During preprocessing, computing the means, variance, and covariance for the data of 7-dimensions took $2.1818$ seconds, that's only a 13\% increase although the data size has increased by 133\%. 

\begin{figure}[!htb]
    \centering
    \begin{subfigure}[t]{0.15\textwidth}
        \centering
        \includegraphics[height=0.9in]{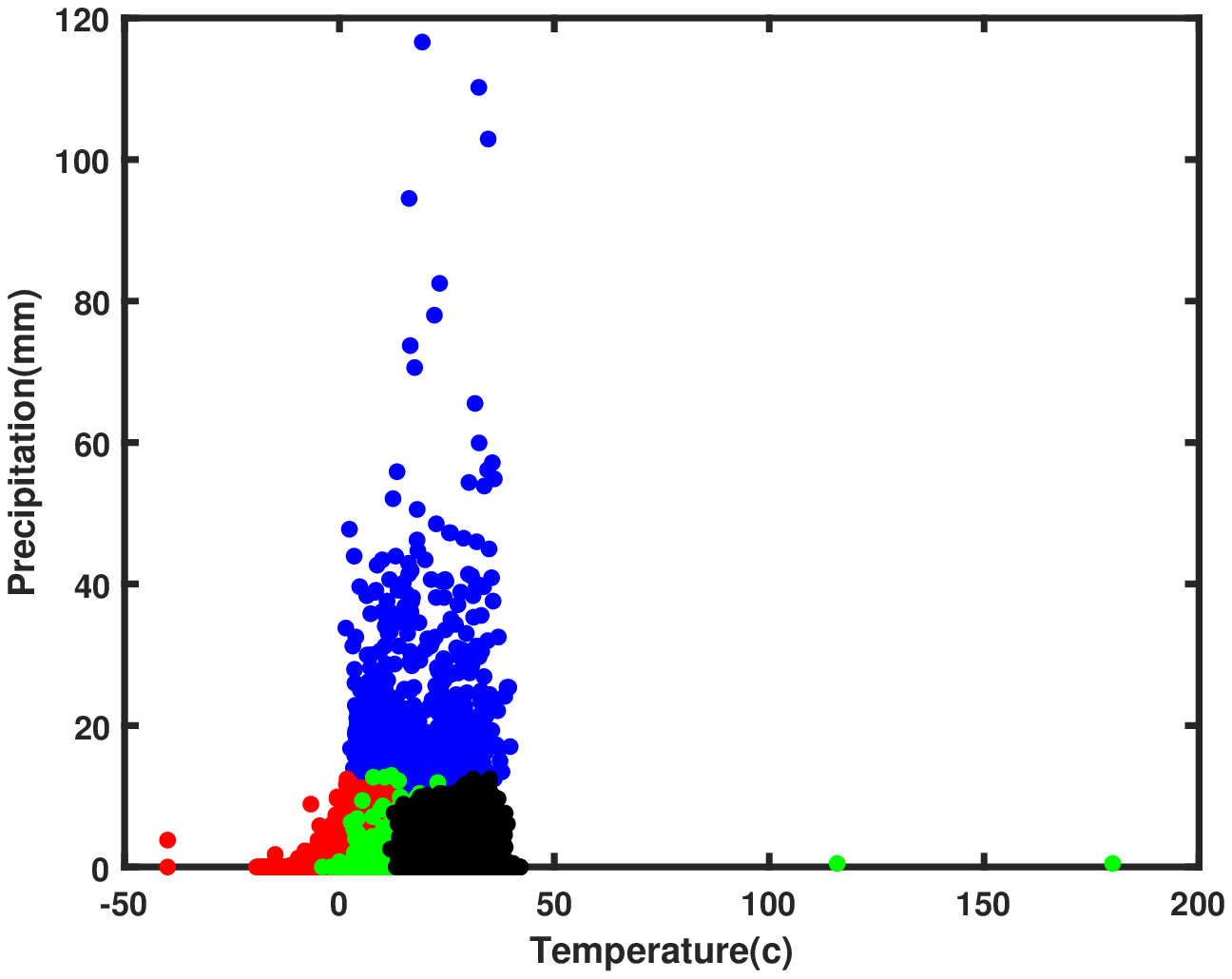}
         \subcaption{}
    \end{subfigure}%
    ~ 
    \begin{subfigure}[t]{0.15\textwidth}
        \centering
        \includegraphics[height=0.9in]{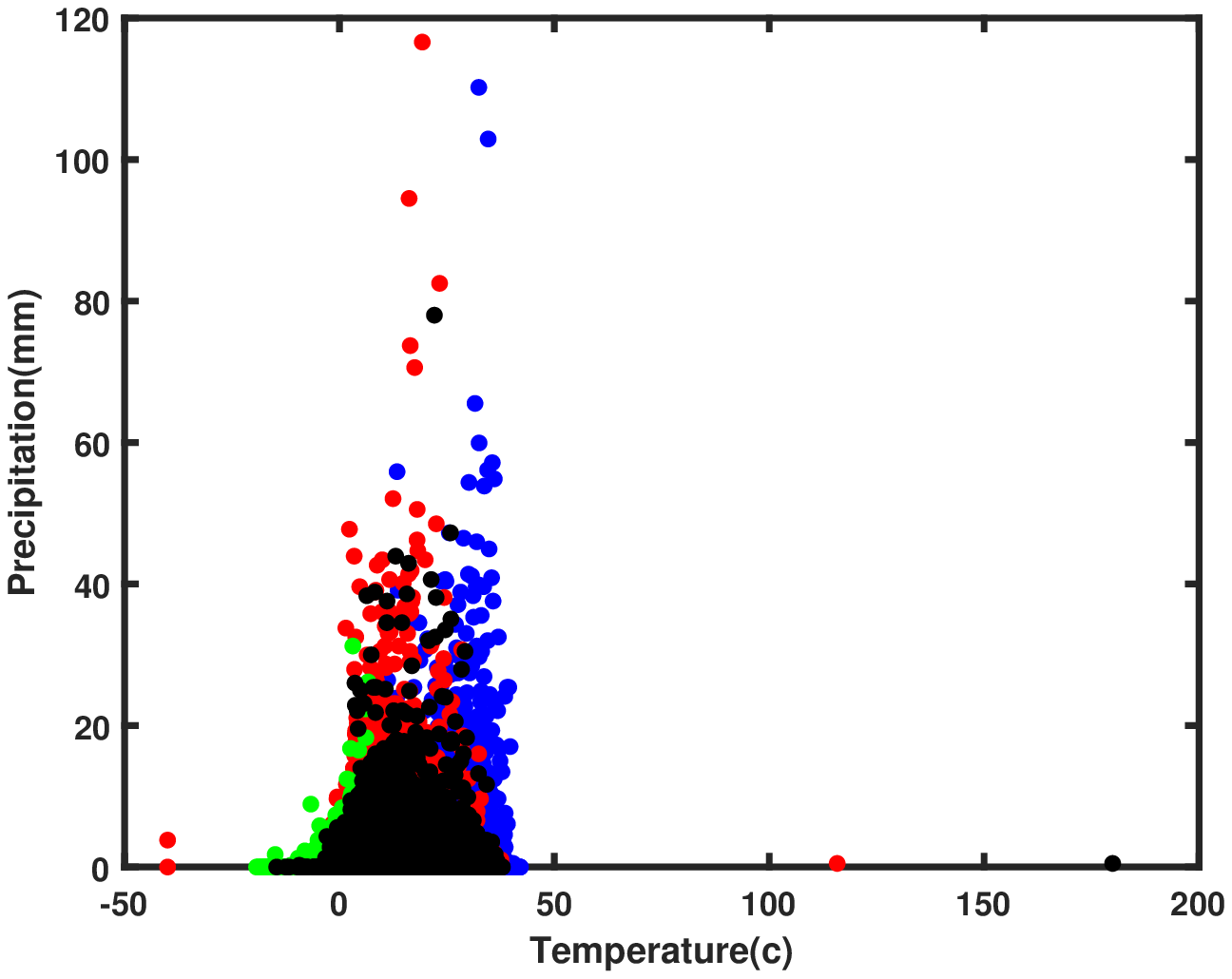}
         \subcaption{}
    \end{subfigure}
    ~ 
     \begin{subfigure}[t]{0.15\textwidth}
        \centering
        \includegraphics[height=0.9in]{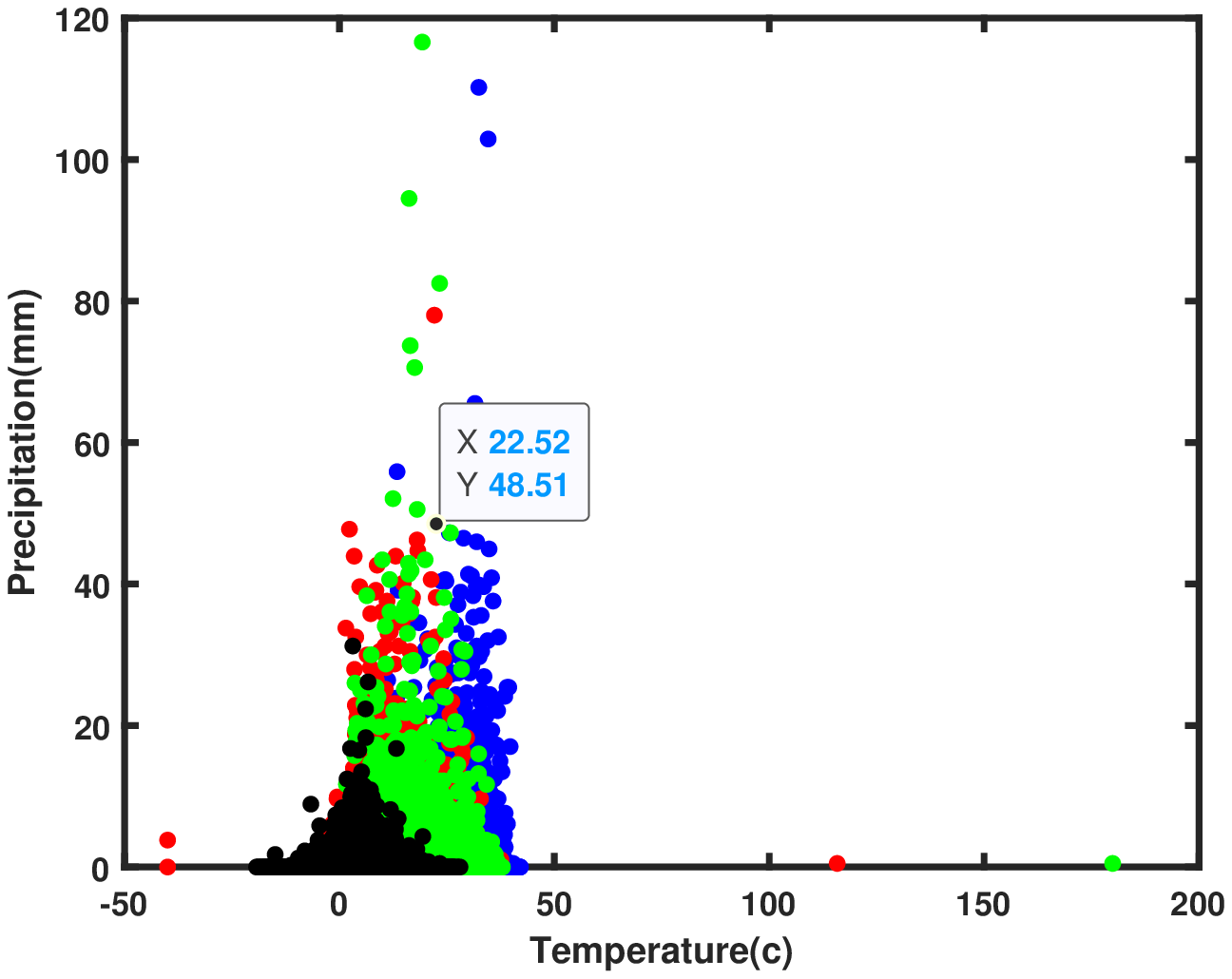}
         \subcaption{}
    \end{subfigure}
   ~
    \begin{subfigure}[t]{0.15\textwidth}
        \centering
        \includegraphics[height=0.9in]{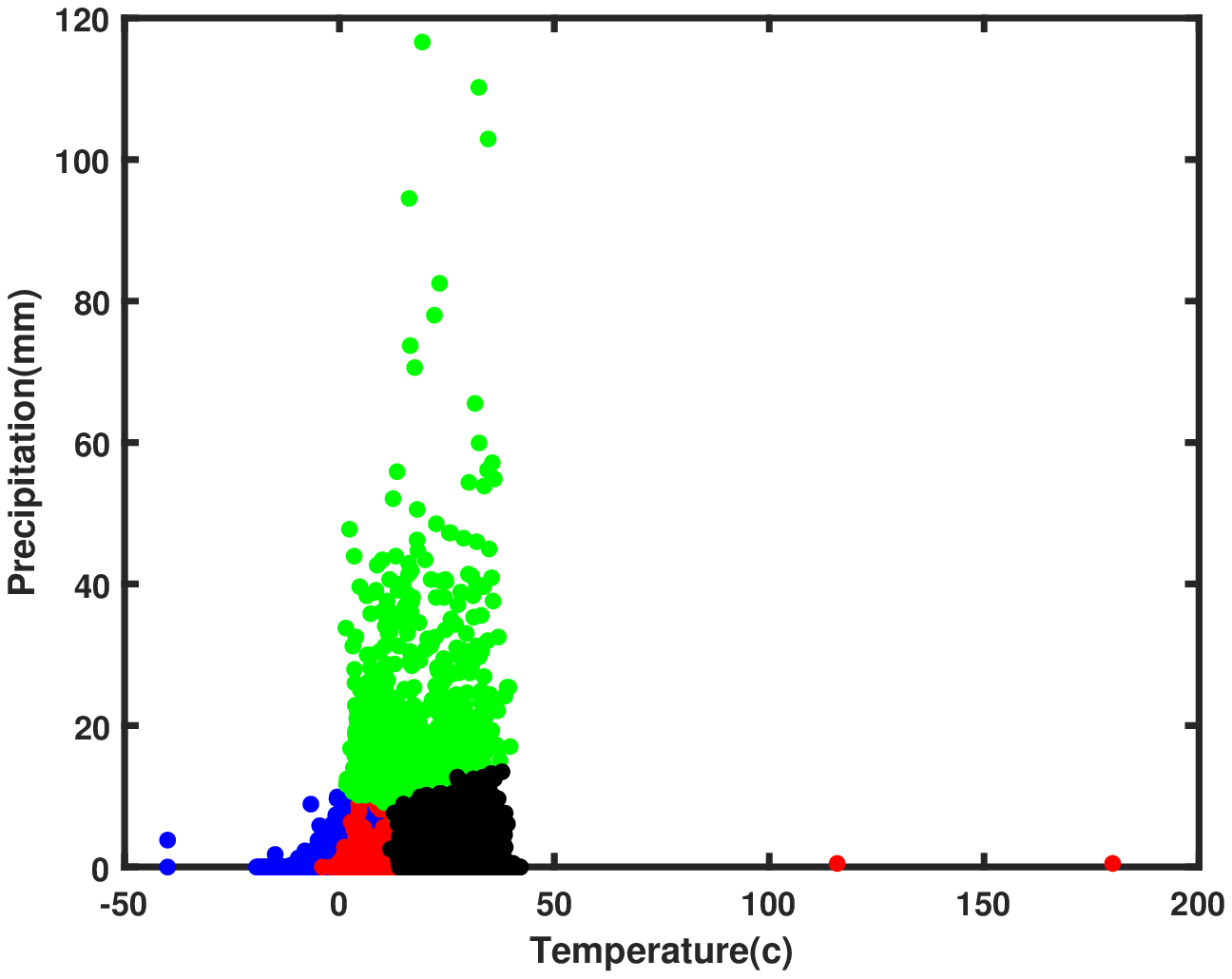}
         \subcaption{}
    \end{subfigure}%
    ~ 
    \begin{subfigure}[t]{0.15\textwidth}
        \centering
        \includegraphics[height=0.9in]{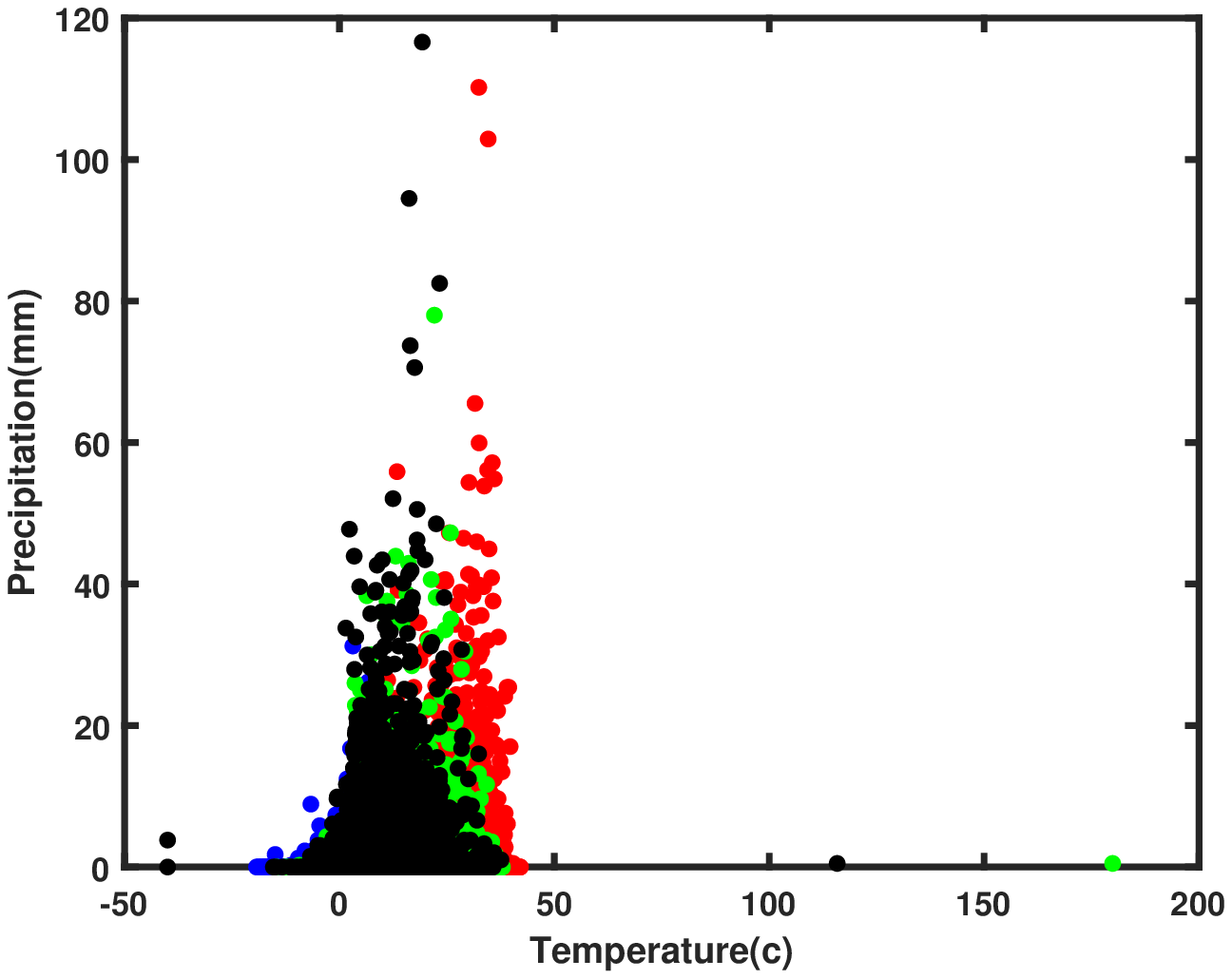}
         \subcaption{}
    \end{subfigure}
    ~ 
     \begin{subfigure}[t]{0.15\textwidth}
        \centering
        \includegraphics[height=0.9in]{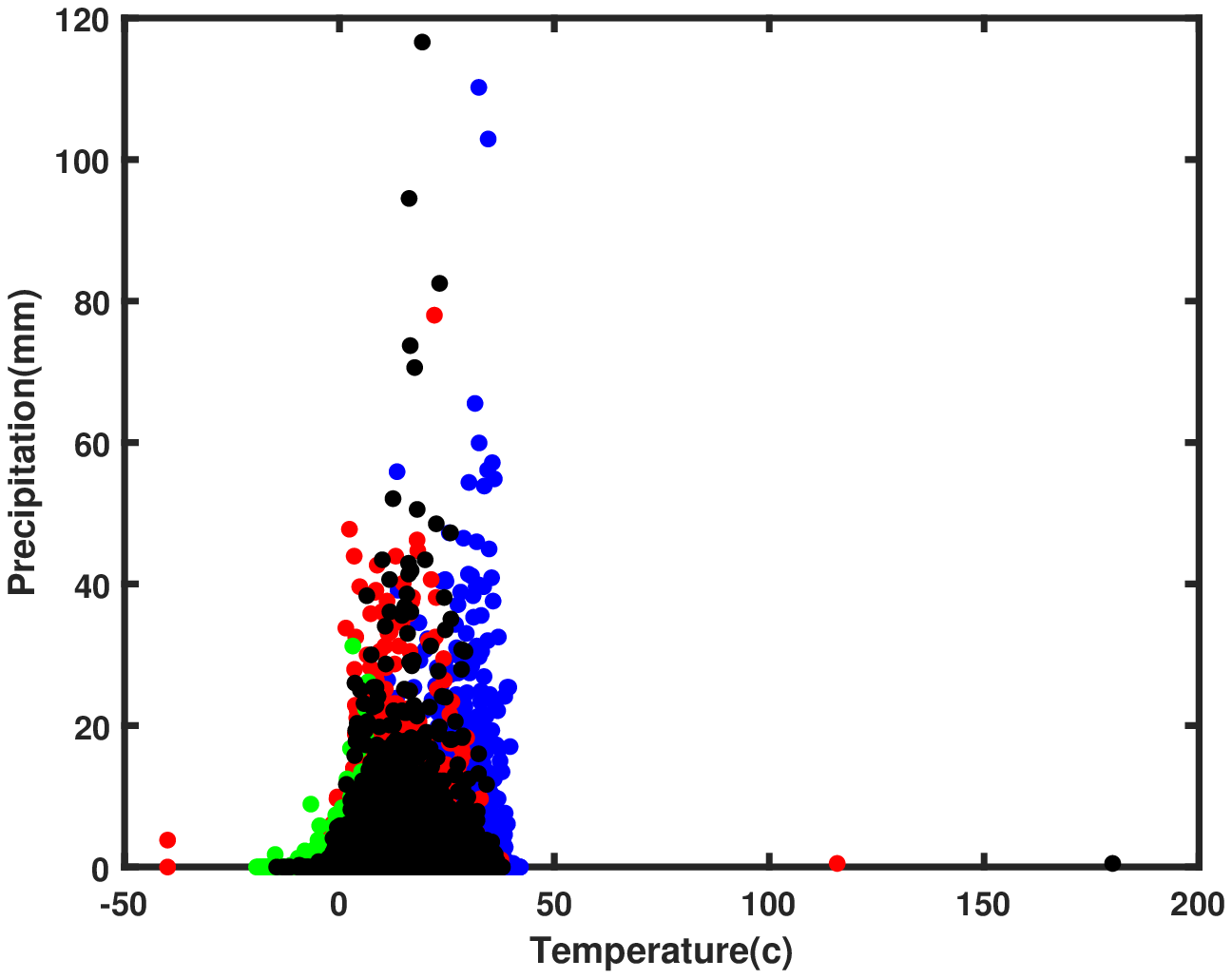}
         \subcaption{}
    \end{subfigure}
    \caption{7-d weather data result: Plots of data clusters (dots)  using  (a) KM, (b) WKM, (c) EKM, (d) KMd, (e) WKMd, (f) EKMd.}
    \label{Cb7}
\end{figure}

The clustering results for $7$-dimensional weather data  with their compute-time  are depicted in Fig.\ref{Cb7} and Table\ref{tab:Cb7}. The summary results show that distribution-based $K$-means and $K$-medoids under both distance measures significantly outperform the corresponding classical versions. The accuracies of KM, WKM, and EKM are  $0.5596$, $0.8595$, and $1.0000$, respectively, while the accuracies of the corresponding $K$-medoids versions are  $0.5578$, $0.8548 $, and $1.0000$, respectively. The corresponding NMIs are $0.7014$, $0.7778$, $1.0000$, $0.2911$, $0.7745$, and $1.0000$, and the corresponding ARIs are $0.6836$, $0.8977$, $1.0000$, $0.7005$, $0.8956$, $1.0000$. The performance progression (classical $< W_2 <$ ED) is consistent across all clustering methods, and the ED-based clustering results offer higher accuracy, NMI, and ARI values than the $W_2$ based ones, plus the compute-time is smaller: As shown in Table~\ref{tab:Cb7}, the overall computational time of our algorithm is less compared to the classical ones that operate on raw data as well as the $W_2$-based ones, yet the accuracy of our algorithms remains higher. This significant performance gain is a result of the use of distributional clustering, where ED outperforms $W_2$, and both outperform the classical methods in terms of accuracy and computation time.

\begin{table}[!htb]
     \centering
      \resizebox{\columnwidth}{!}{                
\begin{tabular}{|c|c|c|c|c|}   
\hline     
 & Accuracy & NMI & ARI& Compute time(s) \\
\hline
KM &$0.5596$ & $0.2941$ & $0.7014$ &$12.6322$\\
\hline                     
WKM &$0.8595 $&$ 0.7778$ & $0.8977$  &$9.6157$  \\
\hline                     
EDKM &$1.0000$ & $1.0000$ &$ 1.0000 $&$8.4273$\\
\hline                     
KMd& $0.5578$ & $0.2911$ & $0.7005$&$5.7702$ \\
\hline                     
WKMd& $0.8548$ &$ 0.7745$ & $0.8956$ &$12.6525$  \\
\hline                     
EKMd& $1.0000$ & $1.0000 $& $1.0000$&$5.7022$ \\
\hline                     
\end{tabular}              
}
\caption{Evaluation measures for 7-D weather data}
 \label{tab:Cb7}
  \end{table}

\subsection{Real-world stocks data}
To demonstrate the effectiveness of our clustering algorithms for also the non-Gaussian distributions, we applied our methods to the stock market price data, which is commonly modeled as lognormal distributions \cite{ANTONIOU2004617}: Letting $X_t$ denote the stock price of a stock $X$ on day $t$, the data $X_t/X_{t-1}$ is assumed to follow lognormal distribution with fixed parameters (that are the means and variances of $\ln(X_t/X_{t-1})$). Manually analyzing and grouping large numbers of stocks with copious data is nearly impossible, but that is necessary for stock analysis. In this regard, automation of clustering methods to group stocks based on their returns is helpful. In this study, we used data from 77 of the top 100 Nasdaq stocks from 2018 to 2019, consisting of $504$ daily adjusted closing prices over the said period of 2 years for each stock. 

We considered each stock as its own random variable, with $504$ days worth of adjusted closing prices to be the supporting dataset. The data size thus equals $77 \times 504=38808$ entries, each of one dimension. For each stock $X$, we examined its $\ln(X_t/X_{t-1})$ values spanning 504 days, and estimated the lognormal parameters described in subsection~\ref{lne} by finding the mean and variance of $\{\ln(X_t/X_{t-1}); 1\leq t\leq 504\}$. We also estimated the covariance of each pair of stocks $X,Y$ by using the data  $\{\ln(X_t/X_{t-1}),\ln(Y_t/Y_{t-1}); 1\leq t\leq 504\}$.  

\begin{figure}[!htb]
\centering
\includegraphics[width= 0.5\textwidth]{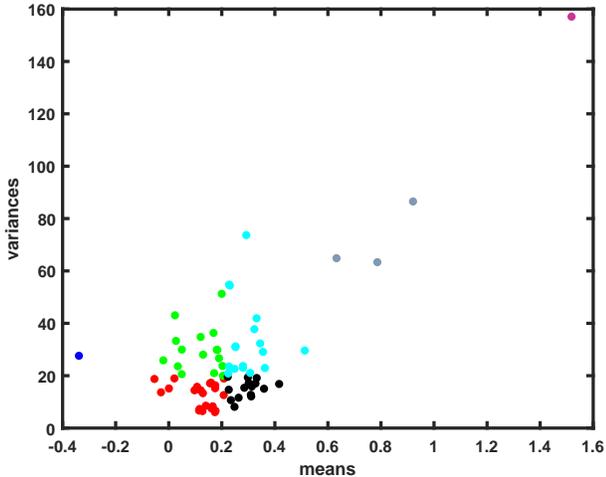}
\caption{Stock data ground clusters} \label{stock}
\end{figure}
    
\begin{table}[!htb]
\centering
 \resizebox{.9\columnwidth}{!}{
\begin{tabular}{|c|r|}
\hline
Class& Stock Label\\
\hline
LR/LV & AEP, AMGN, BKNG, CHTR, CMCSA, CPT,\\
&CSCO, CSX, CTSH, EBAY, EXC, GILD,\\
&GOOG, GOOG, HON, MAR,MDLZ, PAYX,\\
&PCAR, PEP, TMUS, WBA, XEL\\ \hline
LR/MV1 (Very-low)  & KHC\\ \hline
LR/MV2 & ADI, AMAT, ATVI, AVGO, BIIB, DLTR,\\
&EA, META, INTC, MCHP, MNST, NVDA,\\
&NXPI, REGN, SWKS, TXN\\ \hline
MR/LV& AAPL, ADP, ANSS, CPRT, CTAS, FISV,\\
&IDXX, INTU, MSFT, ODFL, ORLY, ROST,\\
&SBUX, SNPS, VRSK, VRSN\\ \hline
MR/MV& ADBE, ADSK, ALGN, AMZN, CDNS, FAST,\\
&FTNT, ILMN, ISRG, KLAC, LRCX, MU,\\
&NFLX, PYPL, QCOM, TSLA, VRTX\\ \hline
HR/MV & AMD, DXCM, MTCH\\ \hline
HR/HV & ENPH\\ 
\hline
\end{tabular}} 
\caption{Ground truth stock clusters from 2018-19~returns}
  \label{Tab:stock}
 \end{table}
 To be able to evaluate the performance (accuracy, NMI, ARI), we created ground truth cluster labels by grouping stocks based on their yearly rate of returns: Low return (LR), Moderate return (MR), High return (HR), Low volatility (LV), Moderate volatility (MV), and High volatility (HV), as documented in Table~\ref{Tab:stock} and clustered in Fig.~\ref{stock}. We used six clustering algorithms, including the classical $K$-means and $K$-medoids and their distribution-based algorithms employing $W_2$ and ED measures. We evaluated the clustering performance using the accuracy, normalized mutual information (NMI), and adjusted Rand index (ARI) metrics.

\begin{table}[!htb]
     \centering
      \resizebox{\columnwidth}{!}{
    \begin{tabular}{|c|c|c|c|c|}
\hline
 & Accuracy & NMI & ARI& Compute time(s) \\
\hline
KM & $0.3766 $&$ 0.1399$ & $0.6391$& $0.523$\\
\hline
WKM& $0.4675 $&$0.4336$  & $0.7468$ & $0.0459$\\
\hline
EKM&$0.5195$&$0.4384$&$0.7488$& $0.1533$ \\
\hline
KMd& $0.3766$  & $0.1362$ & $0.6343$&$0.1015$\\
\hline
WKMd&$0.4935$&$0.4571$&$0.7478$&$ 0.0420$\\
\hline
EKMd &$0.5325$&$0.4651$&$0.7869$&$0.0358$\\
\hline
    \end{tabular} }
    \caption{Evaluation measures for stocks data}
    \label{tab:stockr}
     \end{table}

\begin{figure}[!htb]
    \centering
    \begin{subfigure}[t]{0.15\textwidth}
        \centering
        \includegraphics[height=0.9in]{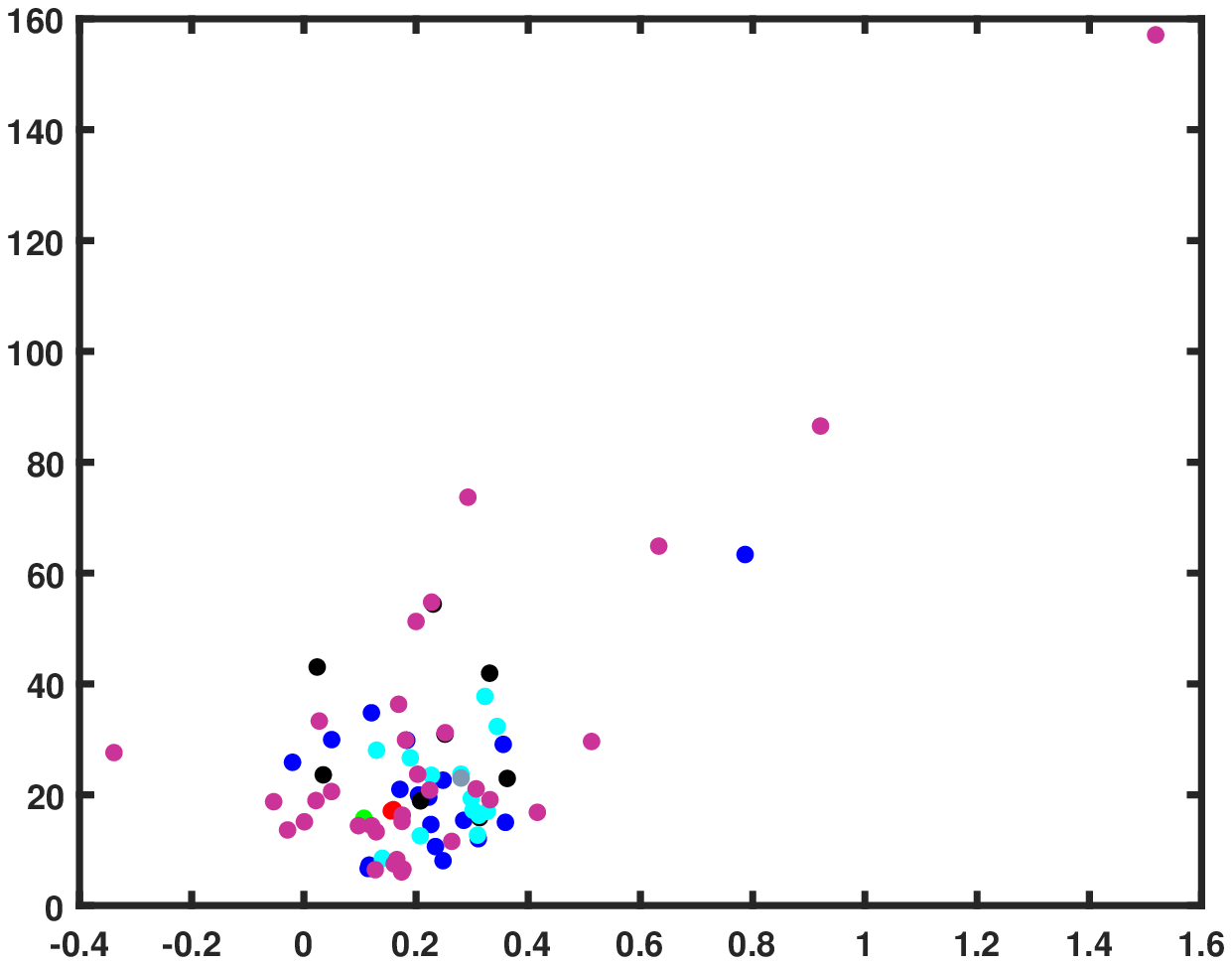}
         \subcaption{}
    \end{subfigure}%
    ~ 
    \begin{subfigure}[t]{0.15\textwidth}
        \centering
        \includegraphics[height=0.9in]{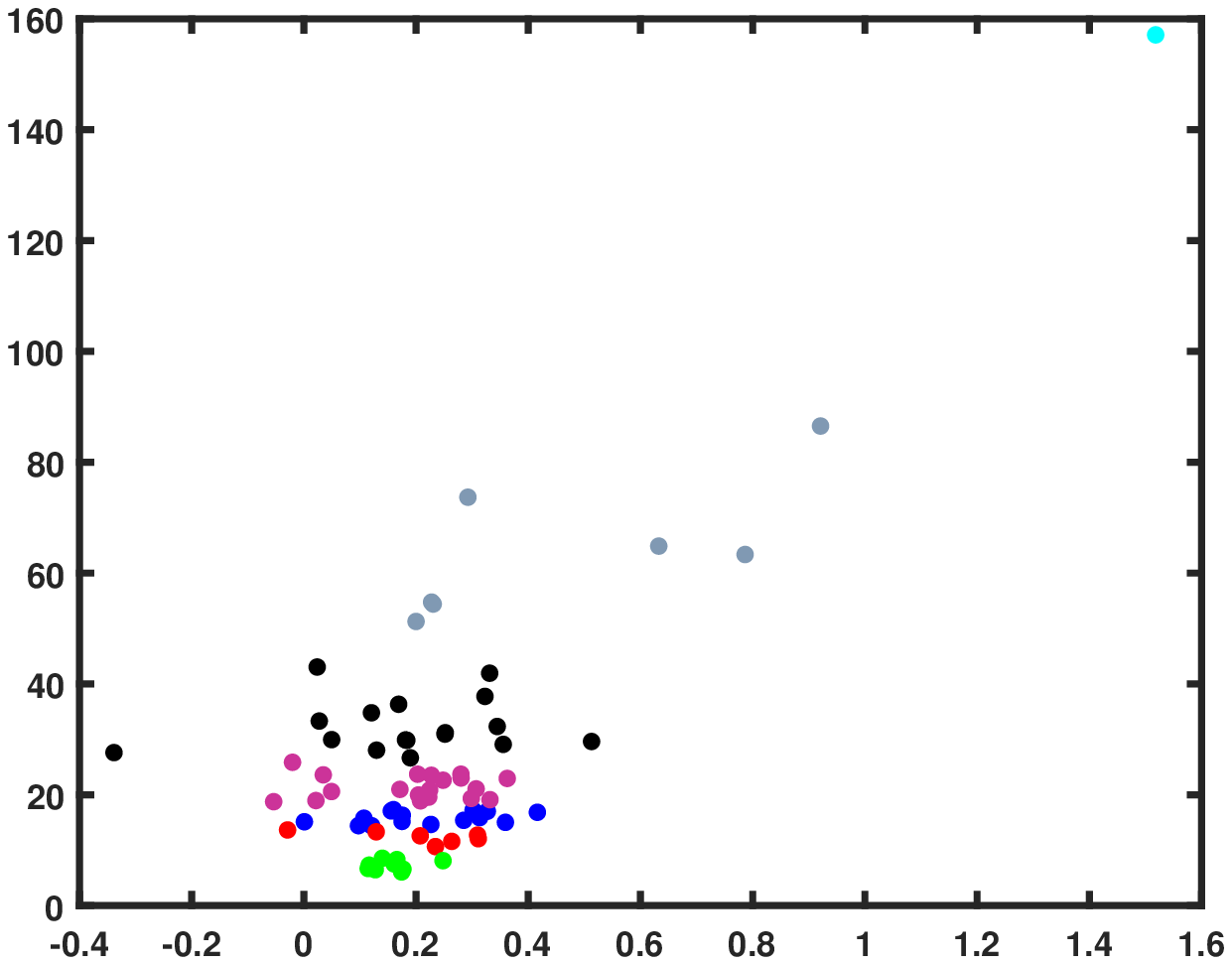}
         \subcaption{}
    \end{subfigure}
    ~ 
     \begin{subfigure}[t]{0.15\textwidth}
        \centering
        \includegraphics[height=0.9in]{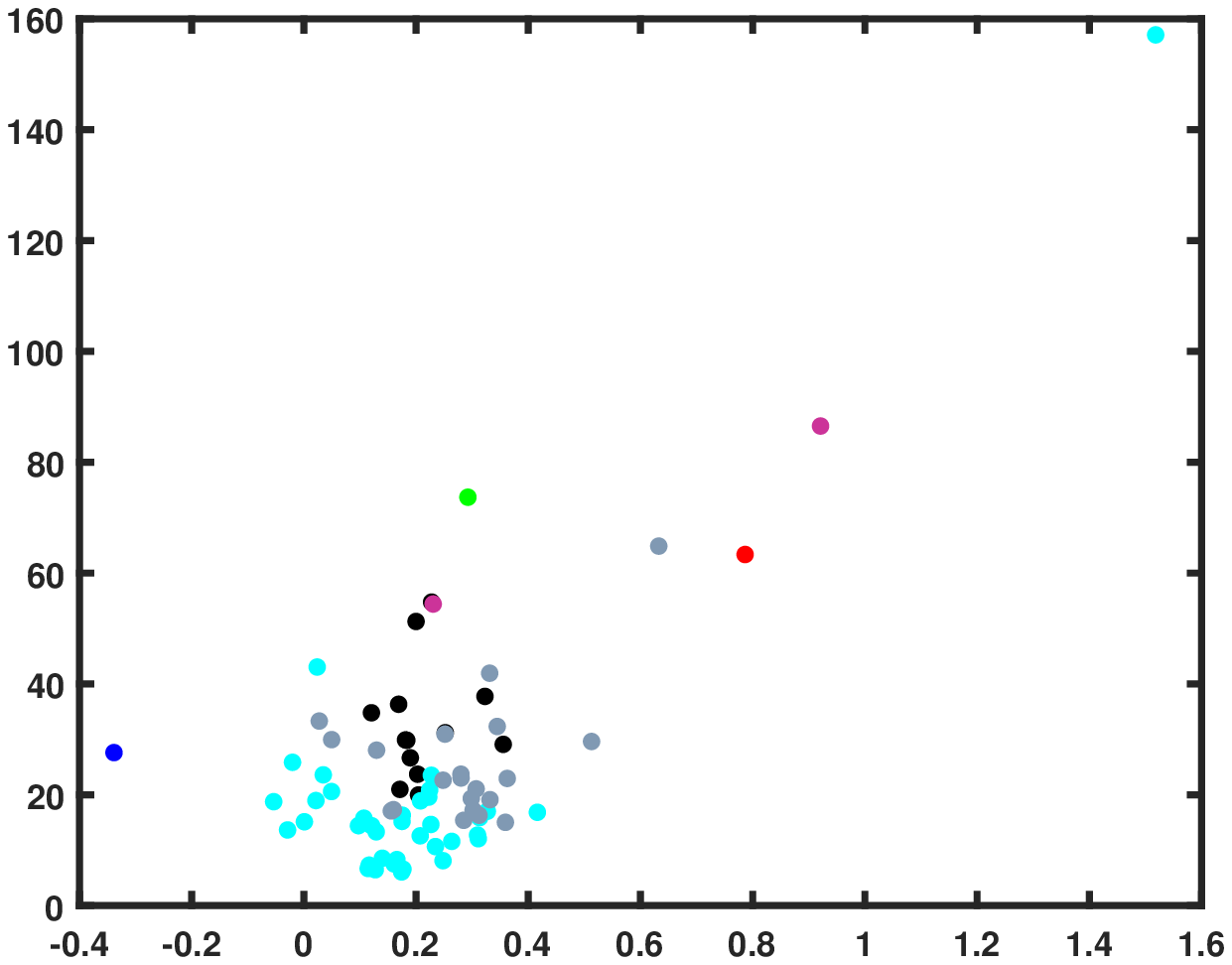}
         \subcaption{}
    \end{subfigure}
   ~
    \begin{subfigure}[t]{0.15\textwidth}
        \centering
        \includegraphics[height=0.9in]{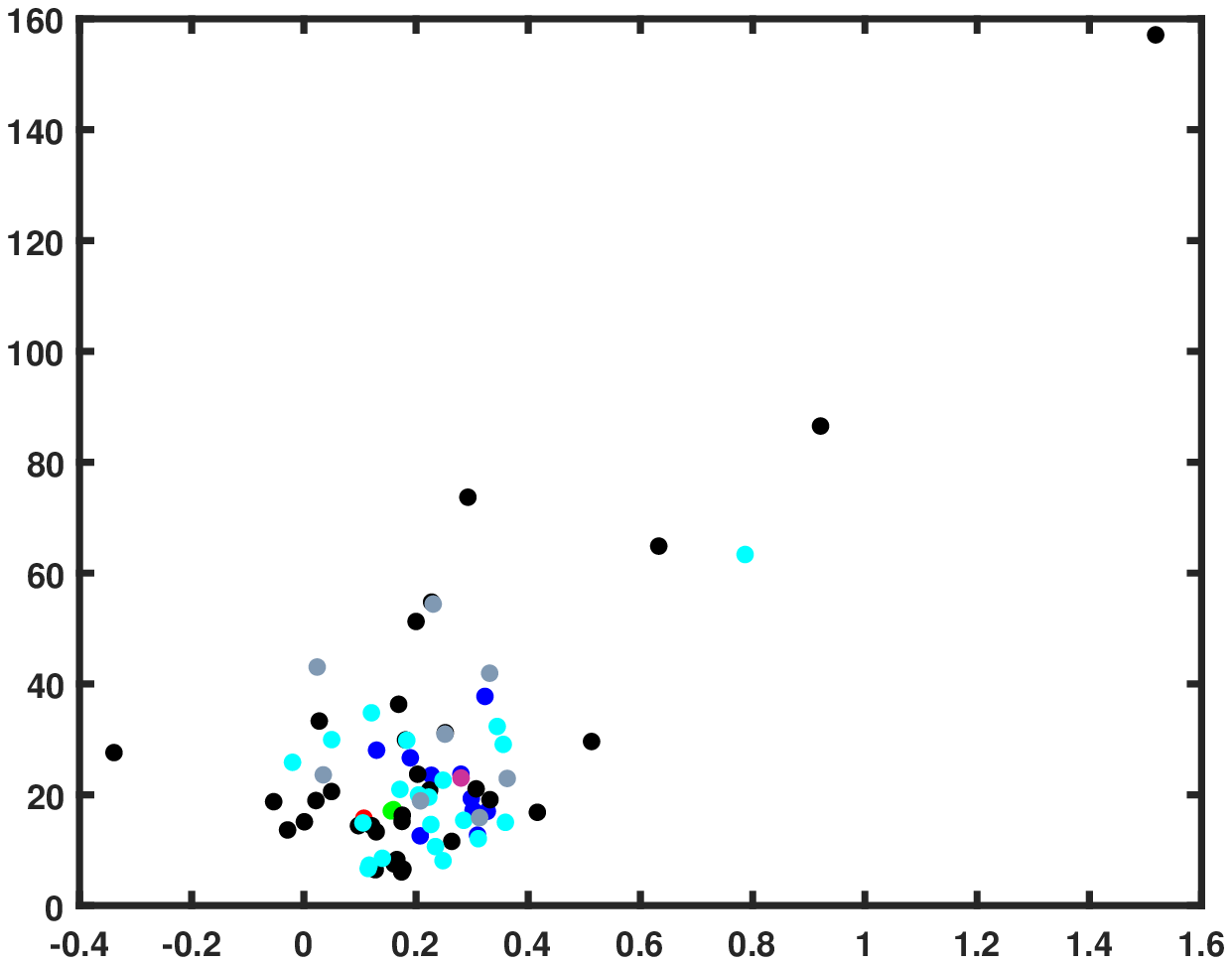}
         \subcaption{}
    \end{subfigure}%
    ~ 
    \begin{subfigure}[t]{0.15\textwidth}
        \centering
        \includegraphics[height=0.9in]{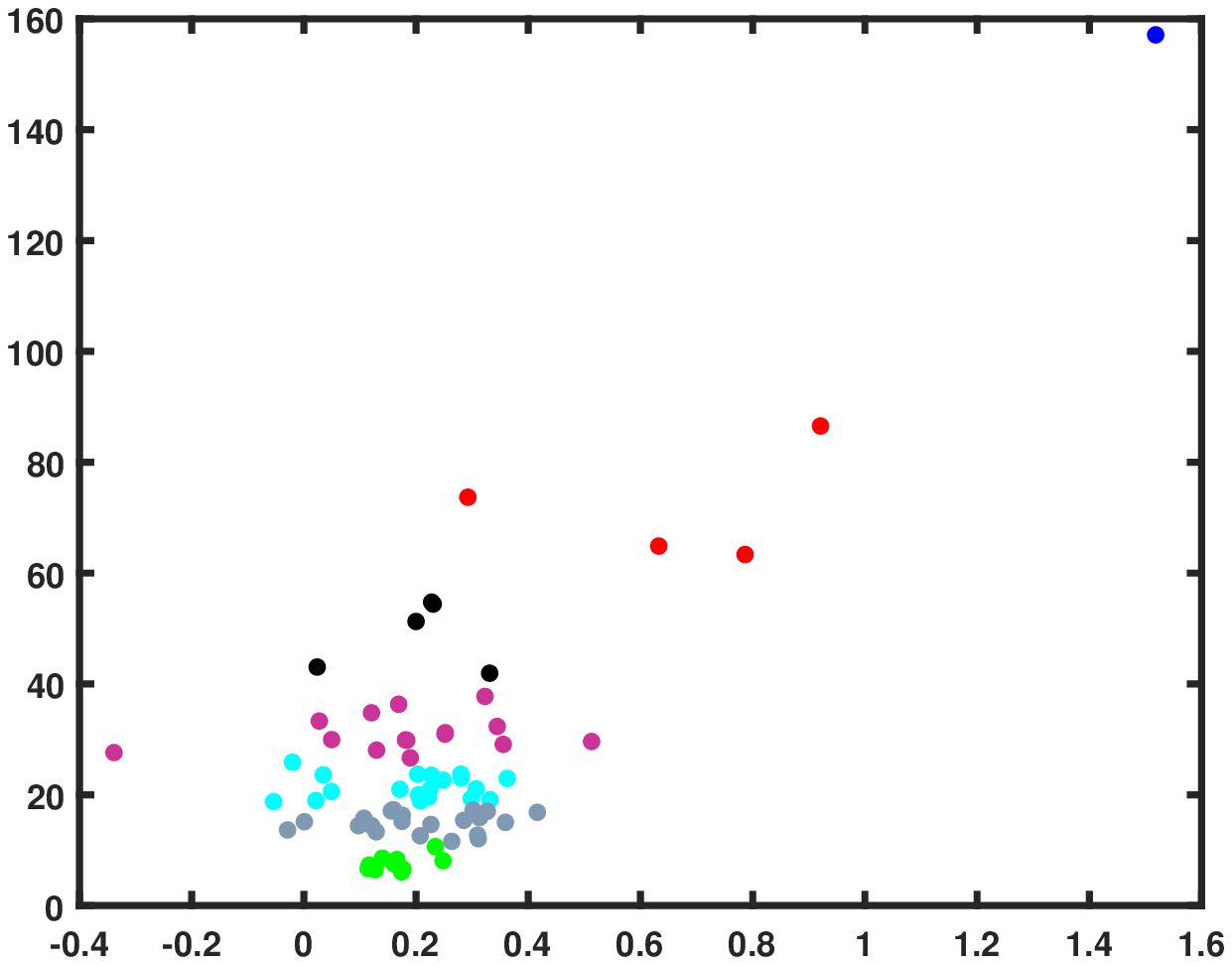}
         \subcaption{}
    \end{subfigure}
    ~ 
     \begin{subfigure}[t]{0.15\textwidth}
        \centering
        \includegraphics[height=0.9in]{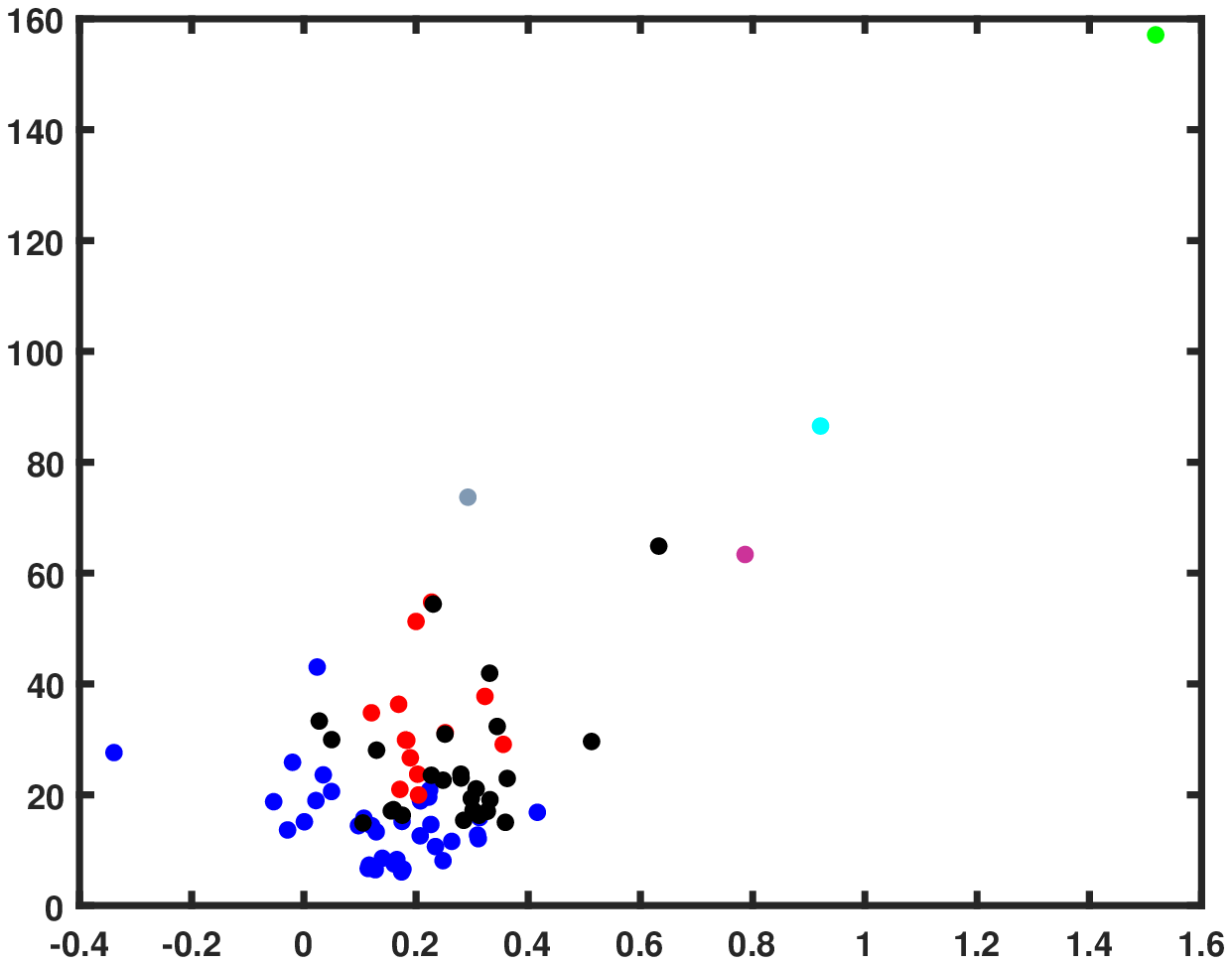}
         \subcaption{}
    \end{subfigure}
    \caption{Stock data result: Plots of data clusters (dots)  using  (a) KM, (b) WKM, (c) EKM, (d) KMd, (e) WKMd, (f) EKMd.}
    \label{fstock}
\end{figure}
The clustering results are shown in Fig~\ref{fstock} and Table~\ref{tab:stockr}. We found that the same progression of performance (classical $< W_2 <$ ED) of the clustering algorithms in terms of accuracy, NMI, ARI, and the compute-time. The accuracies of KM, WKM, and EKM are $0.3766$, $0.4675$, and $0.5195$ respectively, while the accuracies of the corresponding $K$-medoids versions are $0.3766 $, $0.4935$, and $0.5325$ respectively. The corresponding NMIs for $K$-means are: $0.1399$, $0.4336$, $0.4384$, and those for $K$-medoids: $0.1362$, $0.4571$, $0.4651$; and the corresponding ARIs for $K$-means are: $0.6391$, $0.7468 $, $0.7488$, and those for $K$-medoids are $ 0.6343$, $0.7478$, $0.7869$. These demonstrate that ED-based clustering results offer highest accuracy, NMI, and ARI values, while take lesser time to compute as compared to the classical deterministic ones (.1533 sec vs. .523 sec for $K$-means and .0358 sec vs .1015 sec for $K$-mediods). 

In summary, we demonstrated the effectiveness of the clustering algorithms for non-Gaussian distributions using the stock market price data, which is commonly modeled as lognormal distribution.  

\section{Conclusion}
The paper introduced a new distance measure over distributions, called Expectation Distance (ED), and used it to develop noise-robust clustering algorithms, $K$-means, and $K$-medoids. A mathematical derivation proved that the proposed distance is a metric, satisfying the required properties of positivity, symmetry, zero if and only if equal, and triangle inequality. The presented distribution-based $K$-means, and $K$-medoids methods cluster the data distributions first and then assign to each raw-data the cluster of its distribution. The ED-based $K$-means and $K$-medoids and $W_2$-distance based $K$-medoids clustering was introduced for the first time. For both $W_2$ and ED, closed-form expressions for distribution distances were derived in terms of means and covariances, and those values were provided for the case of Gaussian and lognormal distributions. The paper also highlighted that the $W_2$-distance depends only on the marginal distributions, ignoring the correlation information.
In contrast, the proposed ED overcomes this limitation by factoring in the correlation information and, in the process, yields higher noise-robust results. We also noted that while the cluster-centers of the distribution-based $K$-means are independent of the distance measure used, the same is not true of $K$-medoids. We implemented these noise-robust distance-based clustering algorithms and applied them to cluster noisy real-world weather and stocks data by efficiently extracting and using the underlying uncertainty information (in terms of parameters of the distributions---Gaussian in case of weather data and lognormal in case of the stocks data). The results on real-life weather data showed striking improvement in performance for $W_2$-distance and ED-based $K$-means and $K$-medoids, and a higher accuracy was observed for ED in both $K$-means and $K$-medoids: For a $35,280$ entries of $3$-D weather data spanning $4$ seasons over 21 years and $5$ stations, the accuracies of classical $K$-means, $W_2$ $K$-means, and ED $K$-means were found to be $0.5649$, $0.8429$, and $1.0000$ respectively, whereas the accuracies of the corresponding $K$-medoids versions were $0.5637$, $0.8500$, and $0.9976$ respectively. A similar performance progression was also obtained for stock data, demonstrating the method's effectiveness for non-Gaussian distributions. 
This performance validates the noise-robustness of the distribution-data-based clustering schemes and the benefits of factoring in the marginal distributions along with the joint distributions. It was also shown that while the distribution-implied clustering offers higher accuracy than the direct clustering of raw-data, strikingly, the former also has a lower time-complexity. Future research can explore application to other distribution types, such as Gaussian mixtures.
\IEEEpeerreviewmaketitle
 \bibliographystyle{IEEEtran}
 \bibliography{references}
\end{document}